# Quadratic Autoencoder (Q-AE) for Low-dose CT Denoising

Fenglei Fan, *Student Member, IEEE*, Hongming Shan, Mannudeep K. Kalra, Ramandeep Singh, Guhan Qian, Matthew Getzin, Yueyang Teng, Juergen Hahn, and Ge Wang, *Fellow, IEEE*

***Abstract*— Inspired by complexity and diversity of biological neurons, our group proposed quadratic neurons by replacing the inner product in current artificial neurons with a quadratic operation on input data, thereby enhancing the capability of an individual neuron. Along this direction, we are motivated to evaluate the power of quadratic neurons in popular network architectures, simulating human-like learning in the form of "quadratic-neuron-based deep learning". Our prior theoretical studies have shown important merits of quadratic neurons and networks in representation, efficiency, and interpretability. In this paper, we use quadratic neurons to construct an encoder-decoder structure, referred as the quadratic autoencoder, and apply it to low-dose CT denoising. The experimental results on the Mayo low-dose CT dataset demonstrate the utility and robustness of quadratic autoencoder in terms of image denoising and model efficiency. To our best knowledge, this is the first time that the deep learning approach is implemented with a new type of neurons and demonstrates a significant potential in the medical imaging field.**

## I. Introduction

Deep learning [1-2] has achieved great successes in many important fields including medical imaging over the past several years [3-6]. Currently, deep learning research is rapidly evolving and expanding in scope. New results emerge constantly, from theoretical research, architectural innovation, to practical applications. It should be emphasized that the network architecture is critical to the overall performance of a deep learning system, a fact that was testified by the history of deep learning research and development. Numerous excellent models, such as autoencoder [7], VGG [8], Res-Net [9], or GAN [10], were developed, and achieved the state-of-the-art performance, even outperforming human experts in a number of significant tasks, which are frequently reported by media.

The biological neural system is the only known system that defines complex intelligent activities. Deep networks have been successful because they simulate biological neural networks, although the artificial neurons are much simpler than the biological counterparts. Thus, further research in the artificial intelligence / machine learning (AI/ML) field will likely benefit from insights of neuroscience. Still, we are not able to understand all aspects of cortical circuitry and their interplays to form the intelligence, which a deep-learning-based AI system attempts to simulate. However, it is understood that the structural differences between deep learning and cortical circuitry are significant. For instance, biological neurons are of much higher diversity and complexity in terms of morphology and physiology. Nevertheless, typical neural networks only employ simplified homogeneous neurons [11]. Whether the neurological diversity and complexity contribute to human-like learning or not seems an inspiring question.

In reference to the diversity and complexity of biological neurons and human-like learning, our group recently proposed quadratic neurons [12-15] by replacing the inner product in popular artificial neurons with a quadratic operation on input data, thereby enhancing the processing capability of an individual neuron. For instance, even a single quadratic neuron can realize any basic logic/fuzzy logic operation such as XOR. Furthermore, we theoretically demonstrated the advantages of quadratic networks in terms of representation and efficiency [15]. Naturally, we are curious if quadratic networks can deliver competitive results in solving real-world problems as our theoretical analyses suggest; and if so, quadratic-neuron-based deep learning should be included in the machine learning armory, to boost human-like learning towards artificial general intelligence.

Autoencoders [16-22] are a class of unsupervised models that have been successfully applied to denoising, feature extraction and generative tasks. Among the autoencoders, denoising autoencoders [16] were developed based on the idea that a good and robust representation can be adaptively learned from corrupted data themselves. Contractive autoencoders [17] add a penalty term corresponding to the Jacobian matrix to encourage invariance of the representation with respect to small variations in the input data. K-sparse autoencoders [18] only keep the $k$-highest activities so as to prevent the use of an overly large number of units in their encoding phase. Similarly, sparsity can also be enforced by minimizing the Kullback-Leibler (KL) divergence or $l_1$-norm. Variational autoencoders [19] are generative models that can map from random noise to meaningful manifolds. It should be noted that the development of autoencoders is still an evolving area. Recently, generative adversarial networks (GAN) have been incorporated into autoencoders [20], also referred to as adversarial autoencoders, with applications to semi-unsupervised classification tasks and unsupervised clustering. K-competitive autoencoders [21] force neurons in the middle layer to compete for the right of response to the input data so that each neuron is specialized. Graph autoencoders [22] use graph regularization so that the local consistency of a low dimensional manifold can be integrated into representation learning.

This work was supported in part by RPI Clark & Crossan Endowed Fund.

Fenglei Fan, Hongming Shan, Guhan Qian, Matthew Getzin, Juergen Hahn, and Ge Wang* (wangg6@rpi.edu) are with Department of Biomedical Engineering, Rensselaer Polytechnic Institute, Troy, NY, USA. Asterisk indicates the corresponding author.

M. K. Kalra and R. Singh are with Department of Radiology, Massachusetts General Hospital, Boston, MA 02114, USA (e-mail: mkalra@mgh.harvard.edu, rsingh17@mgh.harvard.edu)

Yueyang Teng is with Sino-Dutch Biomedical and Information Engineering School, Northeastern University, Shenyang, China, 110169.

As the first step to prototype a quadratic network, here we propose a quadratic autoencoder that is fundamentally different from the aforementioned autoencoders by virtue of its quadratic feature representation. By definition, our quadratic autoencoder is equipped with quadratic neurons, enhancing the way of how features are extracted and represented. Heuristically, quadratic neurons are important, since the physical world is largely described by second order equations, the geometric objects can be well approximated by splines, and the information world is full of quadratic features. It is therefore hypothesized that a quadratic autoencoder offers the potential to represent complex data more effectively and efficiently than the popular artificial neurons can. To test this hypothesis in the case of low-dose CT denoising, we construct a convolutional quadratic autoencoder and evaluate its performance systematically.

To put our contributions in perspective, we would like to mention that there are relevant results in the literature on high-order or nonlinear representations, either implicitly or explicitly. For example, high-order neurons were used for artificial intelligence [50]. However, they were never connected into a deep network due to training difficulties. On the other hand, our quadratic network gives a high-order nonlinear sparse representation with a reasonable model complexity. Hence, our deep quadratic network can be well trained using modern optimization techniques for practical use. Livni et al. [23] altered the activation function of common artificial neurons to a quadratic form: $\sigma(z) = z^2$, but such a nonlinearity is only of limited utility because at the cellular level the decision boundary is still linear. Tsapanos et al. [24] suggested a parabolic neuron, which is a special case of a quadratic neuron. Lin et al. [25] proposed the so-called network in network (NIN) by sliding micro-networks across an input image so that complex nonlinear feature maps can be acquired. Similarly, Georgios, et al. replaced micro-networks with polynomial kernels [26]. Wang et al. [27] highlighted benefits of second-order operations in machine learning, using the cross product of two network branches, but their method merely produced *de facto* features in the late stage extracted by two fully connected layers consisting of conventional neurons. In contrast to these results, what we propose here is unique in terms of the quadratic neuron structure, the deep quadratic network, and the first of its kind quadratic deep autoencoder.

Low-dose CT denoising has been a long-standing problem in the CT field. Although CT provides critical clinical information, there are potential risks (possible cancer diseases and genetic damages) induced by X-ray radiation [28]. However, reducing radiation dose introduces noise and artifacts in the reconstructed images. To cope with such image degradation, a plethora of algorithms have been developed, which can be classified into the three categories: (i) sinogram filtering [29-31] (ii) iterative reconstruction [32-35], and (iii) image post-processing. Sinogram filtering methods process a sinogram before image reconstruction. However, directly operating on sinogram data may lead to unintended additional artifacts in reconstructed images. Iterative reconstruction algorithms perform image reconstruction iteratively and have gained a popularity over the past decade. However, these algorithms are computationally expensive. Generally speaking, these iterative methods incorporate prior knowledge on data noise and image content into objective functions to reconstruct tomographic images optimally. Different from the first two types of denoising techniques, image post-processing methods directly work on reconstructed low-dose CT images. These post-processing methods have become important for denoising in recent years due to the power of deep learning. Albeit clear improvements in certain areas, traditional image post-processing techniques such as non-local means [36], k-SVD [37], block-matching 3D [38] are often subject to over-smoothness and structural distortion. In contrast, deep learning techniques have delivered a superior performance for low-dose CT denoising. As the first algorithm to apply deep learning to low-dose CT, Chen et al. [39] used a convolutional neural network with 10 layers for denoising. This paper started a fast development of this area, involving the similarity loss, perceptual loss, adversarial mechanism, wavelet, framelet, and transfer learning, as well as various network structures, published as RED-CNN [40], GAN-3D [41], WGAN-VGG [42], CPCE-2D [43], wavelet residual networks [44], SMGAN-3D [45] and so on.

In summary, the contributions of this paper are three-folds:

1) We present a novel and general model to solve the low-dose CT denoising problem, instead of directly applying an existing deep learning technique. To our best knowledge, this is the first paper that uses a deep network consisting of quadratic neurons in the medical imaging field.

2) Our experiments favorably showcase a state-of-the-art denoising performance of the proposed quadratic autoencoder in comparison with other state-of-the-art methods.

3) The quadratic autoencoder enjoys a high model efficiency while achieving a competitive performance, which demonstrates the practical utility of our quadratic autoencoder (Q-AE).

## II. Quadratic Neurons

### *A. Quadratic Neuron*

A quadratic neuron is an upgraded version of a conventional neuron that summarizes input data as an inner product. Mathematically, the quadratic neuron processes the $n$-dimensional input vector, $\boldsymbol{x} = (x_1, x_2, \ldots, x_n)$ in the following manner (other forms of quadratic processing are possible but beyond the scope of this paper):

$$h(\boldsymbol{x}) = (\sum_{i=1}^{n} w_{ir}x_i + b_r)(\sum_{i=1}^{n} w_{ig}x_i + b_g) + \sum_{i=1}^{n} w_{ib}x_i^2 + c$$
$$= (\boldsymbol{w_r}\boldsymbol{x^T} + b_r)\left(\boldsymbol{w_g}\boldsymbol{x^T} + b_g\right) + \boldsymbol{w_b}(\boldsymbol{x^2})^T + c, \quad (1)$$

where $\boldsymbol{x^2}$ denotes point-wise operations. The function $h(\boldsymbol{x})$ will be passed to a nonlinear activation function to define the output:

$$g(\boldsymbol{x}) = \sigma\left(h(\boldsymbol{x})\right), \quad (2)$$

where $g(\boldsymbol{x})$ expresses a typical quadratic neuron with $\sigma(\cdot)$ as an activation function such as a rectified linear unit (ReLU). It is noted that our definition of a quadratic neuron only uses $3n$ parameters, which is much sparser than the general quadratic representation demanding $\frac{n(n+1)}{2}$ parameters. In [26], the quadratic filter is explored of the complexity $O(n^2)$, which is more challenging to use in a deep network due to the combinatorial explosion in training cost and risk of over-fitting. By the way, the parabolic neuron presented in [27] can be directly included as a special case of our definition.

### B. "Quadratic Convolution"

Convolutional neural networks (CNN) are the most significant architecture for deep learning. Specifically, for quadratic networks, each feature map is obtained by sliding a quadratic neuron over an input field in the same manner as a micro network does in the context of NIN [25]. Strictly speaking, such a quadratic operation is not a linear convolution, but it can be regarded as a nonlinear version of the conventional convolution.

Suppose that $\boldsymbol{x}$ is an input image, and the conventional convolution operation is denoted as $*$, the computational model in a conventional neuron is:

$$m_1(\boldsymbol{x}) = \sigma(\boldsymbol{W} * \boldsymbol{x} + b), \tag{3}$$

In contrast, the feature map of a quadratic neuron is given by:

$$m_2(\boldsymbol{x}) = \sigma((\boldsymbol{W_r} * \boldsymbol{x} + b_r)(\boldsymbol{W_g} * \boldsymbol{x} + b_g) + \boldsymbol{W_b} * \boldsymbol{x^2} + c), \tag{4}$$

where $b_r, b_g, c$ are biases and $\boldsymbol{W_r}, \boldsymbol{W_g}, \boldsymbol{W_b}$ are 2D kernels.

It is principally correct that given sufficiently many conventional neurons, a conventional network can approximate a network consisting of quadratic neurons. A more sensible question is that if one quadratic neuron is always equivalent to composition/addition of three or a few conventional neurons, and a negative answer would immediately suggest a value of a quadratic neuron. Actually, this is not the case. For example, suppose that $\sigma(\cdot)$ is a piecewise linear function ReLU, then common operations on conventional neurons only yield a piecewise linear function instead of a nonlinear function. Therefore, it can be argued that the quadratic neuron is an extension to that conventional neuron and cannot be always simulated by a number of conventional neurons. Also, it is emphasized that a quadratic neuron is not just using a quadratic activation, and actually the quadratic neuron can use the activation function of the conventional neuron directly. The quadratic neuron is characterized by the employment of a quadratic processing operation prior to activation.

### C. Algebraic Structure and Model Efficiency

A more detailed discussion on theoretical properties of quadratic networks is warranted. In this subsection, let us offer several insights for peace of mind. Most of these results are from our earlier analyses.

**Algebraic Structure:** A quadratic network with depth of $O(\log_2(N))$ and width of no more than $N$ can represent any univariate polynomial of degree $N$ [14].

The proof is based on Algebraic Fundamental Theorem: without involving complex numbers, any univariate polynomial of degree $N$ can be factorized as $P_N(x) = C\prod_i^{l_1}(x - x_i)\prod_j^{l_2}(x^2 + a_j x + b_j)$, where $l_1 + 2l_2 = N$. This theorem indicates indispensability and uniqueness of a quadratic expression. One question that naturally arises about the necessity of quadratic neurons is why quadratic neurons are focused on, instead of third-order or fourth-order neurons. With the Algebraic Fundamental Theorem, it is clear that higher order representations are not as fundamental as the first order and quadratic. Since the space of polynomial functions is dense in the space of differential functions [51], it can be justified that the quadratic expression is basic and complementary to the linear counterpart.

**Efficiency Theorem:** Given the network with only one hidden layer, there exists a function that a quadratic network can approximate with a polynomial number of neurons while a conventional network can only do the same level approximation with exponentially more neurons [14].

The complexity of a neural network is approximately given by the product of the complexity of the structure multiplied by the complexity of the neuron as the building block. By this theorem, in spite of a mildly increased neural complexity ($O(3n)$ over $O(n)$), the structural complexity of the quadratic network can be significantly smaller than its conventional counterpart due to the need for a smaller number of quadratic neurons. An experimental demonstration of this key point is given below.

### D. Training Quadratic Networks

Training a deep neural network is to optimize the involved parameters by minimizing its loss function. In the same spirit, training a quadratic deep network is in principle just like training a conventional one. The optimization of a neural network repeats the layer-wise chain-rule-guided backpropagation until convergence. To adjust the weights of quadratic neurons, we need to know output of each layer $y(\boldsymbol{x})$ and compute the gradients with respect to the weights (and biases) of the quadratic neurons. The gradient terms will be used to update the parameters and minimize the loss. The key formulas are as follows:

$$\begin{cases} \dfrac{\partial y}{\partial \boldsymbol{W_r^i}} = \boldsymbol{x^i}(\boldsymbol{W_g} * \boldsymbol{x} + b_g) \\ \dfrac{\partial y}{\partial b_r} = \boldsymbol{W_g} * \boldsymbol{x} + b_g \\ \dfrac{\partial y}{\partial \boldsymbol{W_g^i}} = \boldsymbol{x^i}(\boldsymbol{W_r} * \boldsymbol{x} + b_r) \\ \dfrac{\partial y}{\partial b_g} = \boldsymbol{W_r} * \boldsymbol{x} + b_r \\ \dfrac{\partial y}{\partial \boldsymbol{W_b^i}} = (\boldsymbol{x^i})^2 \\ \dfrac{\partial y}{\partial c} = 1 \\ \dfrac{\partial y}{\partial \boldsymbol{x^i}} = \boldsymbol{W_g^i}(\boldsymbol{W_r} * \boldsymbol{x} + b_r) + \boldsymbol{W_r^i}(\boldsymbol{W_g} * \boldsymbol{x} + b_g) + 2\boldsymbol{W_r^i}\boldsymbol{x^i} \end{cases} \tag{5}$$

### E. General Autoencoder Model

A general autoencoder model with regularization terms imposed on both network parameters and latent features can be expressed as follows:

$$\underset{\boldsymbol{W_E},\boldsymbol{W_D}}{\operatorname{argmin}} \left\|\boldsymbol{Y} - \boldsymbol{\mathcal{H}_D}\big(\boldsymbol{W_D}, \boldsymbol{\mathcal{H}_E}(\boldsymbol{W_E}, \boldsymbol{X})\big)\right\|_2^2 + R_1\big(\boldsymbol{\mathcal{H}_E}(\boldsymbol{W_E}, \boldsymbol{X})\big) + R_2(\boldsymbol{W_E}, \boldsymbol{W_D}) \tag{6}$$

where $\boldsymbol{X}$ is the input, $\boldsymbol{Y}$ is the label in supervised tasks. In the case of unsupervised learning, $\boldsymbol{Y}$ is $\boldsymbol{X}$ itself. $\boldsymbol{\mathcal{H}_E}$ is an encoding function mapping the input to latent features, $\boldsymbol{\mathcal{H}_D}$ is a decoding function, $\boldsymbol{W_E}$ and $\boldsymbol{W_D}$ are encoding and decoding parameters, and $R_1$ and $R_2$ are regularization functions acting upon features and weights respectively. For example, $R_1$ can express sparsity and/or invariance constraints, and $R_2$ plays a role to prevent overfitting, such as in terms of the $l_1$-norm. In this study, no regularization terms were used, and we set $R_1 = 0, R_2 = 0$. That is, we purposely focus on the generic quadratic autoencoder and its intrinsic potential so that any gain from the quadratic network cannot be attributed to regularization.

## III. Low-Dose CT Experiment

### *A. Network Design*

Overall, the performance of a neural network is determined by four factors: network architecture, loss function, optimization strategy, and dataset. The network architecture controls its intrinsic representation ability. Generally speaking, the larger the model is, the more powerful its representation ability is. The loss function measures the network performance according to some metrics, with or without regularization techniques. The optimization strategy focuses on how to adjust the parameters and achieve the learning goal. Last but not least, the quality and size of the dataset is particularly important for data-driven deep learning. It is commonly believed that the landscape of the loss function includes a large number of local minima. The Adam [46] algorithm produces excellent results in terms of training speed and overall performance, without being trapped at local minima. In our experiment, we will contrast our methods with other deep learning algorithms, such as CNN10, RED-CNN, WGAN-VGG and CPCE-2D, as summarized in TABLE I.

**Network architecture:** The key elements of various autoencoders are convolutional/deconvolutional layers, shortcut connections, bottleneck layers, and zero padding schemes. Deconvolutional operations, which refers to the transposed convolutional operations, are inverses of convolutional operations. Convolution generally diminishes the size of an input, being undesirable for denoising models that keep the size of a noised image after denoising. Therefore, zero padding is necessary. The bottleneck layer is a latent space of the auto-encoding workflow, where noise and redundant information is removed. However, essential information may also be discarded in this process. To compensate for such a loss, the bypass connections are imposed to reuse earlier features, hence retaining structural details and improving spatial resolution. The bypass connection can be in one of the two primary forms: a residual shortcut as the identical mapping like that in ResNet [8] and a concatenate of earlier feature maps with latter ones as used in DenseNet [47].

**Objective function:** Various objective functions represent different penalizing mechanisms to produce a learned mapping. Although it is reported that minimizing the MSE between the denoised and normal dose CT images may yield an over-smooth appearance, it is possible to avoid such a degradation by controlling model complexity and training time. The adversarial loss (AL) in the GAN is an alternative way that can learn the distribution from low-dose to normal-dose CT images but may not necessarily provide the true image content in an individual case. The perceptual loss (PL) imitates the human vision system in perceiving images by adapting an ImageNet-based pre-trained VGG model. However, the perceptual loss performs inferiorly in terms of noise removal. The similarity loss (SL) can be regarded as a surrogate of PL since the rationale for SL is to preserve structural and textual information as well. Again, in our network design we use the common loss function MSE to evaluate the intrinsic representation ability of the proposed quadratic autoencoder against the other models. Clearly, it is possible to integrate various loss functions into our quadratic autoencoder.

TABLE I: COMPARISON BETWEEN REPRESENTATIVE DENOISING MODELS

| Methods | Conv | De-conv | Shortcut | Zero-Padding | Bottleneck Layer | Objective function |
|---|---|---|---|---|---|---|
| CNN10 | ✓ | ✗ | ✗ | ✓ | ✗ | MSE |
| RED-CNN | ✓ | ✓ | Residual | ✗ | ✓ | MSE |
| WGAN-VGG | ✓ | ✗ | ✗ | ✓ | ✗ | AL+PL |
| CPCE-2D | ✓ | ✓ | Concatenation | ✗ | ✓ | AL+PL |
| SMGAN-3D | ✓ | ✗ | ✗ | ✓ | ✗ | AL+SL |
| Q-AE(Ours) | ✓ | ✓ | Residual | ✓ | ✓ | MSE |

### *B. Our Methodology*

#### *1) Quadratic Autoencoder Using Residual Shortcuts:*

In [16], the noise was added to the input images before they were fed into the autoencoder. Therefore, the autoencoder is appropriate for restoring the original signals from degraded copies in a supervised fashion. For image restoration, a fully connected autoencoder is not effective because of its inability to extract local features from image manifolds. In contrast, we employ quadratic convolutional and deconvolutional operations for low-dose CT denoising. In this context, to enhance model trainability and features reuse, residual connections are resorted in the form of symmetric bridging between the convolutional and deconvolutional layers. Like in [40], ReLU operations are performed for activation in our networks.

Overall, the structure of the proposed quadratic autoencoder employs quadratic filters and traverses the gulf between paired convolutional and deconvolutional layers. The topology of our network architecture is shown in Figure 1. There are 5 quadratic convolutional layers and 5 quadratic deconvolutional layers in our Q-AE, where each layer has 15 quadratic filters of $5 \times 5$, followed by a ReLU layer. Zero paddings are used in the first four layer, with the fifth layer as the bottleneck layer. Although we draw on the experience from RED-CNN [40] in terms of the employment of deconvolutional layers and the arrangement of residual shortcuts, our model is different from RED-CNN because Q-AE is totally constructed with quadratic neurons. The Q-AE components are detailed in the following:

**Quadratic Filter:** As illustrated earlier, the quadratic filter is more powerful than the linear filter which is focused on linear features. As indicated by our earlier theorems, learning nonlinear features with quadratic features has the potential to improve the network capability (such as an improved denoising performance) and/or reduce the model complexity (with a reduced number of network parameters).

**Encoder-Decoder Structure:** We integrate stacked convolutional layers into an encoder-decoder structure. In order to preserve local correlation in the encoding phase, we do not utilize pooling layers and also apply zero-padding operations in the first four convolutional layers. The fifth convolutional layer serves as the bottleneck layer without zero padding. Encouraged by their successes in biomedical image applications, we incorporate deconvolutional layers into our stacked decoder for the sake of preserving structural details. Because the encoder and decoder are symmetric, the deconvolutional layers should be brought in alignment with the corresponding convolutional layers. To this end, we ensure that the kernel sizes of the convolutional and deconvolutional layers match exactly. Also, the first deconvolutional layer is not zero-padded while other deconvolutional layers are zero-padded.

**Residual Shortcuts:** The strengths of residual shortcuts are two-fold. First, the use of shortcut connections can solve the training difficulty with deep models. Feed forward neural networks are not good at learning the identical mapping [8], and the residual shortcuts can help avoid the gradient explosion/vanishing when training a deep network. Second, feature reuse helps to preserve high-resolution structural and contrast details, which can greatly boost the network performance [8]. Motivated by the success of the RED-CNN network, we arrange the positions of three shortcuts in the same way as shown in Figure 1. In the decoding phase, the ReLU were used as the activation function after the summation with the residual mapping.

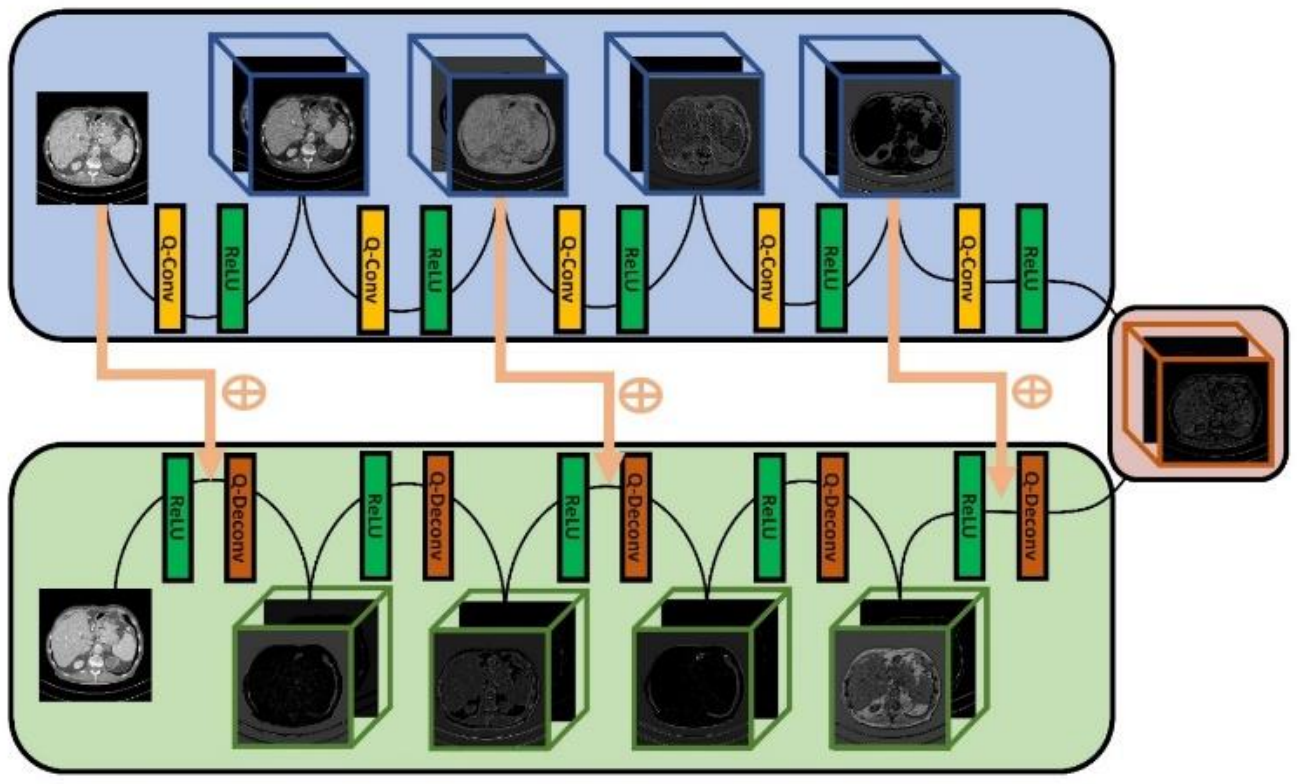

Figure 1. Architecture of the proposed quadratic autoencoder.

### *2). Experimental Setting*

We randomly extracted 74,000 64*64 patches from the five patient datasets in the Mayo low-dose CT challenge dataset. In this set of patches, 64,000 were used as training dataset, and the rest 10,000 as validation dataset. All the images were normalized to [0,1] from the window [-300, 300] HU. A mini-batch size of 50 was used for each iteration, with 20 epochs for all the deep learning methods. The Adam optimization was applied for training in the TensorFlow.

In the experiments, we selected the algorithms CNN10 [39], WGAN-VGG [42], RED-CNN [40], CPCE-2D [43] and BM3D [38] as the baselines in comparison with our quadratic autoencoder. All these algorithms are well known in the CT field. Among them, while BM3D is a classic algorithm, the others are deep learning methods dedicated to low-dose CT. For all deep learning models, we use mean squared error as loss function, which is generally effective in denoising because we want to fairly evaluate their intrinsic performance under the same conditions. Originally, RED-CNN has phenomenally over-smooth denoising results. It is feasible to circumvent this problem by slimming the network. Meanwhile, to keep the reasonable model complexity, we reduced 96 filters to 32 filters in each layer.

To conduct an unbiased comparison, we should use optimal hyperparameters for the competing algorithms. For the hyper-parameters in WGAN-VGG, we used the configuration recommended in [42]. For CNN10, RED-CNN and CPCE-2D, the most influential parameter is the learning rate for the Adam optimization. Here we determined the best rate based on the minimum final validation loss from one random initialization. The possible learning rate set chosen for CNN10, RED-CNN and CPCE-2D are $\{1\times 10^{-4},\ 2\times 10^{-4},\ 5\times 10^{-4}\ 8\times 10^{-4},\ 1.0\times 10^{-3},\ 1.2\times 10^{-3},\ 1.5\times 10^{-3}\}$. The learning rate values are plotted with the corresponding validation loss values in Figure 2. It can be seen that the lowest validation errors are obtained at $5\times 10^{-4}$, $5\times 10^{-4}$ and $5\times 10^{-4}$ for CNN10, RED-CNN, and CPCE-2D respectively. Therefore, we use those learning rates that gave the lowest errors for the comparison, as summarized in TABLE II. As far as the Q-AE is concerned, for training $\boldsymbol{w_r}$ and $\boldsymbol{w_g}$ of each layer were randomly initialized with a truncated Gaussian function, $b_g$ are set to 1 for all the layers. In this way, quadratic term $(\boldsymbol{w_r x^T} + b_r)(\boldsymbol{w_g x^T} + b_g)$ turns into linear term $(\boldsymbol{w_r x^T} + b_r)$. The reason why we use such initialization is because quadratic terms should not be pre-determined, they should be learned in the training. $b_r$ and $c$ set to 0 initially for all the layers. $\boldsymbol{w_b}$ was set to 0 here, we will discuss the influence of $\boldsymbol{w_b}$ on the network in the context of direct initialization and transfer learning later. The learning rate was set to $4\times 10^{-4}$ for the first 10 epochs, and $2\times 10^{-4}$ for the rest epochs. A total number of epochs is 30 to guarantee the convergence of all models.

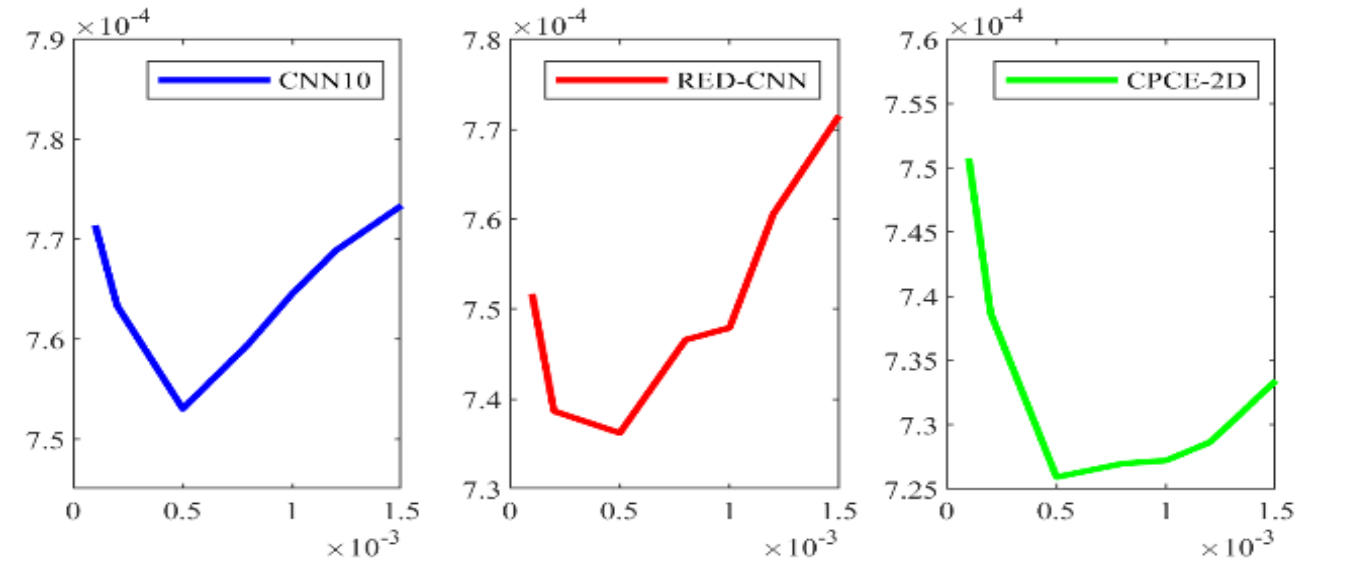

Fig. 2. Losses versus learning rates for CNN10, RED-CNN and CPCE-2D respectively.

TABLE II: PARAMETER SETTINGS FOR FOUR competing METHODS.

| Methods | Revision | Parameter configuration | Parameters |
|---|---|---|---|
| **CNN10** | ✗ | Optimization | s. t. $=8\times 10^{-4}$ |
| **RED-CNN** | sliming | Optimization | s. t. $= 4\times 10^{-4}$ |
| **WGAN-VGG** | ✗ | Inheritance [42] | Inheritance [42] |
| **CPCE-2D** | loss replacement | Optimization | s. t.$= 8\times 10^{-4}$ |

Note: s.t. = study rate.

TABLE III: QUANTITATIVE COMPARISON OF THE SELECTED DENOISING ALGORITHMS.

| | | **LDCT** | **Q-AE** | **CNN10** | **CPCE-2D** | **BM3D** | **RED-CNN** | **WGAN-VGG** |
|---|---|---|---|---|---|---|---|---|
| L286 | PSNR | 30.430±0.951 | 32.619±1.242 | 32.366±1.221 | 32.559±1.299 | 31.160±2.064 | 32.674±1.271 | 30.430±0.951 |
| | SSIM | 0.916±0.020 | 0.935±0.015 | 0.933±0.015 | 0.934±0.015 | 0.838±0.057 | 0.935±0.015 | 0.910±0.020 |
| | RMSE | 0.0303±0.003 | 0.0236±0.003 | 0.0243±0.003 | 0.0238±0.003 | 0.0279±0.005 | 0.0235±0.003 | 0.0303±0.003 |
| L291 | PSNR | 28.476±1.802 | 30.202±2.085 | 30.060±2.001 | 30.220±2.042 | 29.018±2.593 | 30.301±2.033 | 28.476±1.802 |
| | SSIM | 0.887±0.042 | 0.905±0.038 | 0.903±0.040 | 0.905±0.039 | 0.823±0.060 | 0.906±0.039 | 0.0887±0.042 |
| | RMSE | 0.0385±0.008 | 0.0318±0.008 | 0.0323±0.008 | 0.0317±0.008 | 0.0370±0.011 | 0.0314±0.007 | 0.0385±0.008 |
| L310 | PSNR | 28.497±1.367 | 30.261±1.514 | 30.075±1.487 | 30.234±1.492 | 29.221±1.908 | 30.318±1.490 | 28.450±1.367 |
| | SSIM | 0.862±0.031 | 0.891±0.026 | 0.889±0.027 | 0.890±0.026 | 0.775±0.050 | 0.891±0.026 | 0.862±0.003 |
| | RMSE | 0.0382±0.006 | 0.0311±0.005 | 0.0318±0.005 | 0.0312±0.005 | 0.0348±0.006 | 0.0309±0.005 | 0.0380±0.006 |
| L333 | PSNR | 29.3780±0.940 | 31.236±1.127 | 30.955±1.126 | 31.120±1.128 | 29.670±1.581 | 31.270±1.126 | 29.380±0.940 |
| | SSIM | 0.908±0.0180 | 0.924±0.018 | 0.922±0.017 | 0.927±0.017 | 0.845±0.044 | 0.924±0.017 | 0.908±0.018 |
| | RMSE | 0.0342±0.004 | 0.0277±0.004 | 0.0286±0.004 | 0.0278±0.004 | 0.0328±0.004 | 0.0276±0.004 | 0.0342±0.004 |
| L506 | PSNR | 30.435±1.579 | 32.328±1.800 | 32.089±1.742 | 32.334±1.781 | 31.190±2.294 | 32.402±1.781 | 30.435±1.579 |
| | SSIM | 0.932±0.026 | 0.942±0.024 | 0.940±0.024 | 0.942±0.024 | 0.886±0.050 | 0.943±0.0234 | 0.932±0.026 |
| | RMSE | 0.0306±0.006 | 0.0248±0.006 | 0.0254±0.006 | 0.0247±0.006 | 0.0280±0.007 | 0.0245±0.006 | 0.0306±0.006 |

### *3). Convergence Behaviors*

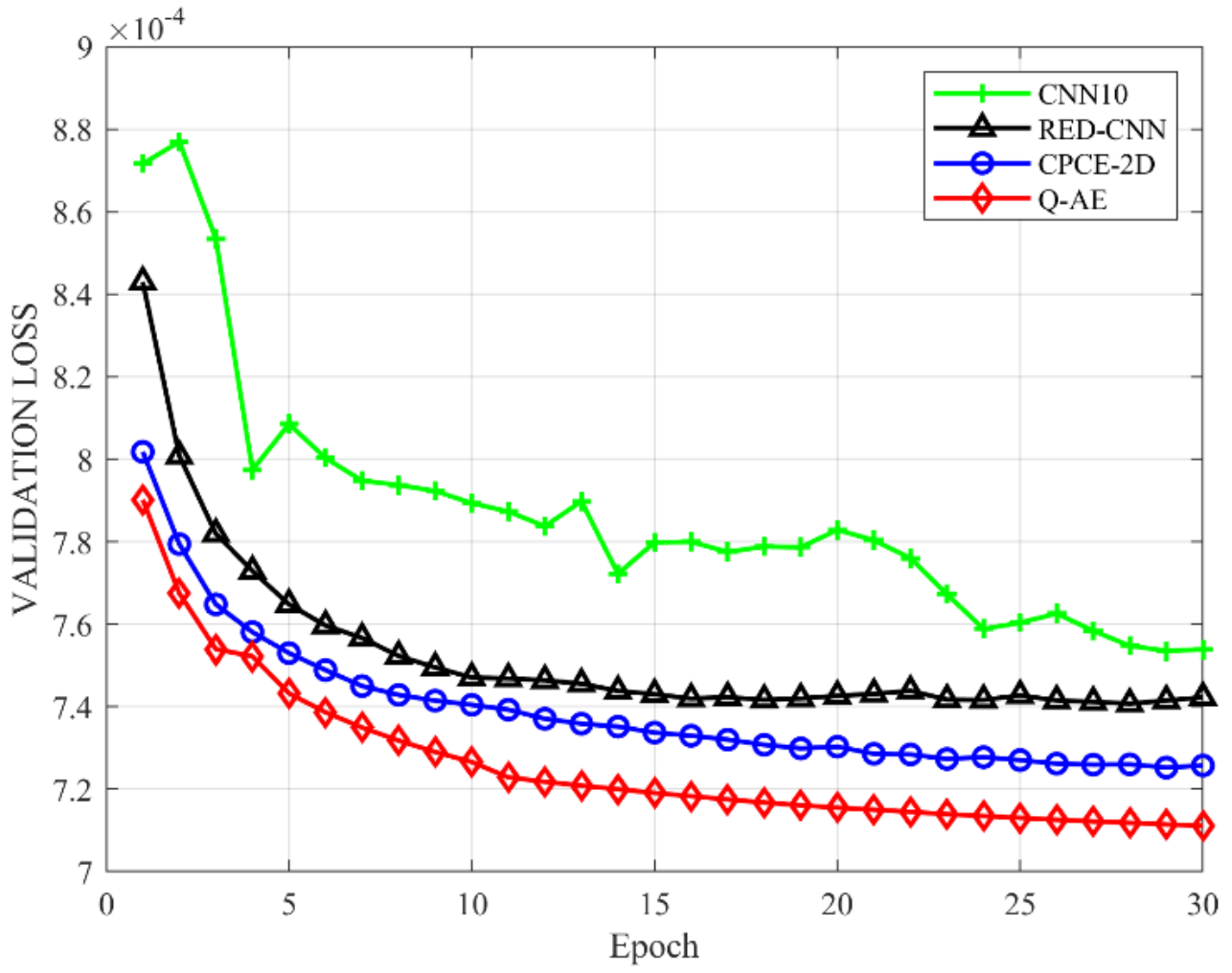

Fig. 3. Comparison of the convergence behaviors of the proposed quadratic autoencoder (Q-AE), CNN10, RED-CNN and CPCE-2D.

The convergence properties of the selected models are compared in Figure 3. The plots in Figure 3 show that the quadratic autoencoder (Q-AE) achieved the lowest loss value, compared to its competitors. The training of the feed forward network CNN10 is challenging. The trajectory of its loss function fluctuates from time to time. In contrast, the trajectory for RED-CNN, CPCE-2D and Q-AE are monotonously decreasing and then being stable. In fact, with the residual shortcuts, deep networks with even thousands of layers are trainable [8]. It should be noted that CPCE-2D and Q-AE started from almost the same level after the first epoch. Then, they kept similar paces with the following four epochs. Noticeably, the downward trend of Q-AE still sustains for late epochs that of CPCE-2D, i.e., the validation loss of Q-AE descends more quickly. Even after only six epochs, Q-AE surpassed the final performance of CPCE-2D, indicating a great time saving in training Q-AE.

### *4). Denoising Performance*

Two representative slices (specifically, 270$^{th}$ slice from patient L506 and 340$^{th}$ slice from patient L310) containing low-attenuation lesions and blood vessels were selected from two patients to visualize the denoising performance as shown in Figures 4 and 6. For better comparison, we zoomed those ROIs that are marked by red rectangles in Figures 5 and 7. Generally speaking, all the networks demonstrate the denoising effect to different degrees. The output images of CNN-10, RED-CNN, CPCE-2D and Q-AE are well denoised but compromised by the smoothening effect due to the use of MSE as the loss function. Comparing Figure 4 with Figure 6 shows that the results of BM3D suffer from streak artifacts, WGAN-VGG shows intended structural fidelity but allows the existence of significant noise, which somehow comprises the SSIM values. The CNN10 results are somewhat blurred and RED-CNN alleviates over-smoothness due to the decreased number of filters. CPCE-2D has a decent denoising performance and preservation of textural information is well.

TABLE IV: MEDIAN SCORES OF READER STUDY FOR SIX ALGORITHMS

| Metrics | Image Texture | | | Noise | | | Fidelity | | |
|---|---|---|---|---|---|---|---|---|---|
| | R1 | R2 | R3 | R1 | R2 | R3 | R1 | R2 | R3 |
| Clean | 3 | 3 | 3 | 3 | 2 | 3 | 2 | 2 | 2 |
| CNN10 | 2 | 4 | 2 | 3 | 3 | 3 | 2 | 3 | 2 |
| CPCE-2D | 2 | 4 | 2 | 3 | 3 | 3 | 2 | 3 | 2 |
| BM3D | 1 | 3 | 1 | 1 | 1 | 1 | 1 | 3 | 1 |
| RED-CNN | 2 | 4 | 2 | 4 | 4 | 4 | 3 | 3 | 3 |
| **Q-AE** | **3** | **4** | **2** | **4** | **4** | **4** | **3** | **3** | **3** |
| WGAN-VGG | 3 | 2 | 2 | 3 | 3 | 3 | 2 | 2 | 2 |
| Noised | 2 | 1 | 1 | 1 | 1 | 1 | 1 | 1 | 1 |

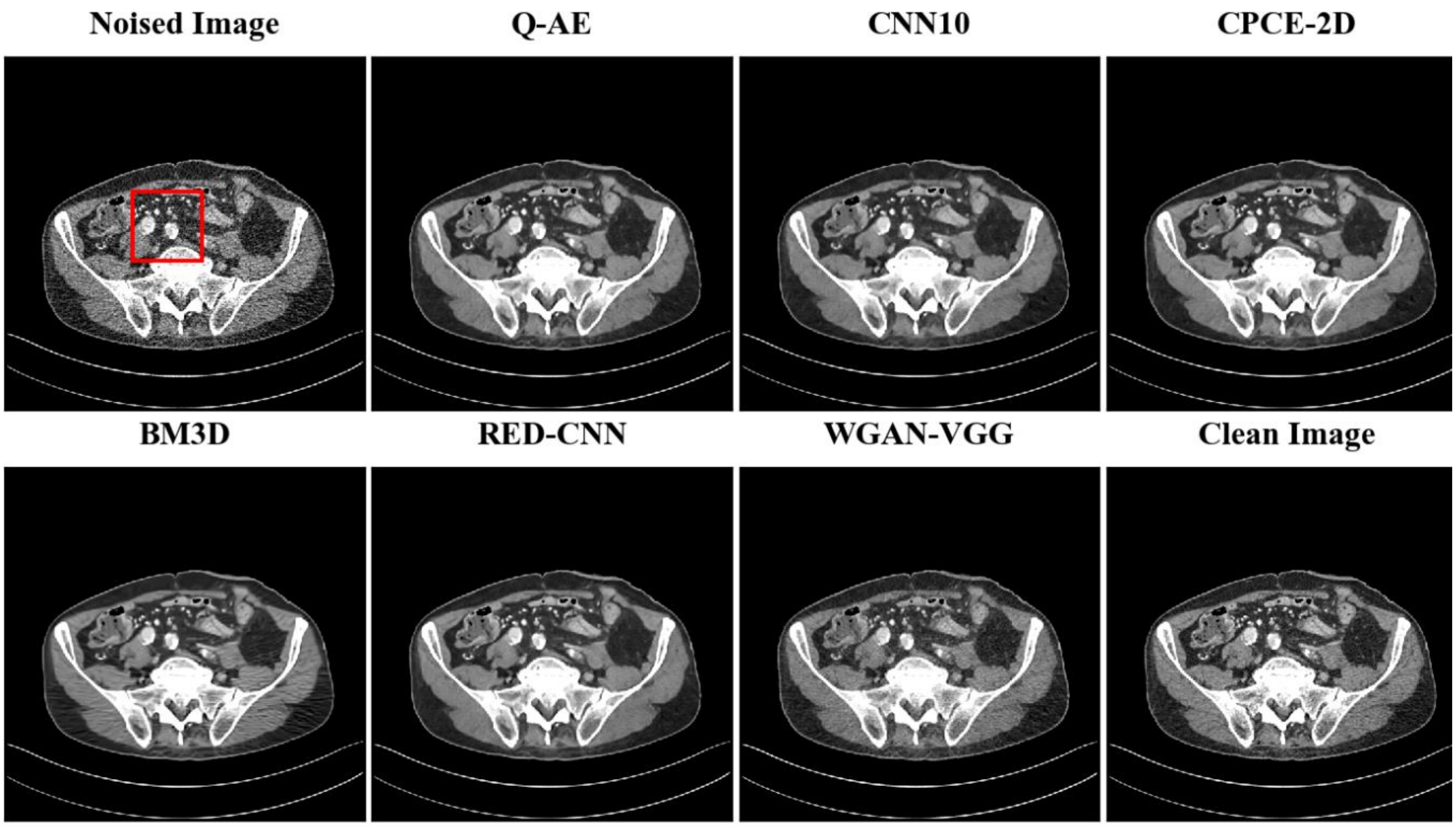

Figure 4. Comparison of different deep learning methods for denoising low dose CT images. Compared to the other techniques, the Q-AE image has lower image noise, maintained conspicuity of small organism (in the small red circle), and superior fidelity of small structures. Display window = [-160, 240].

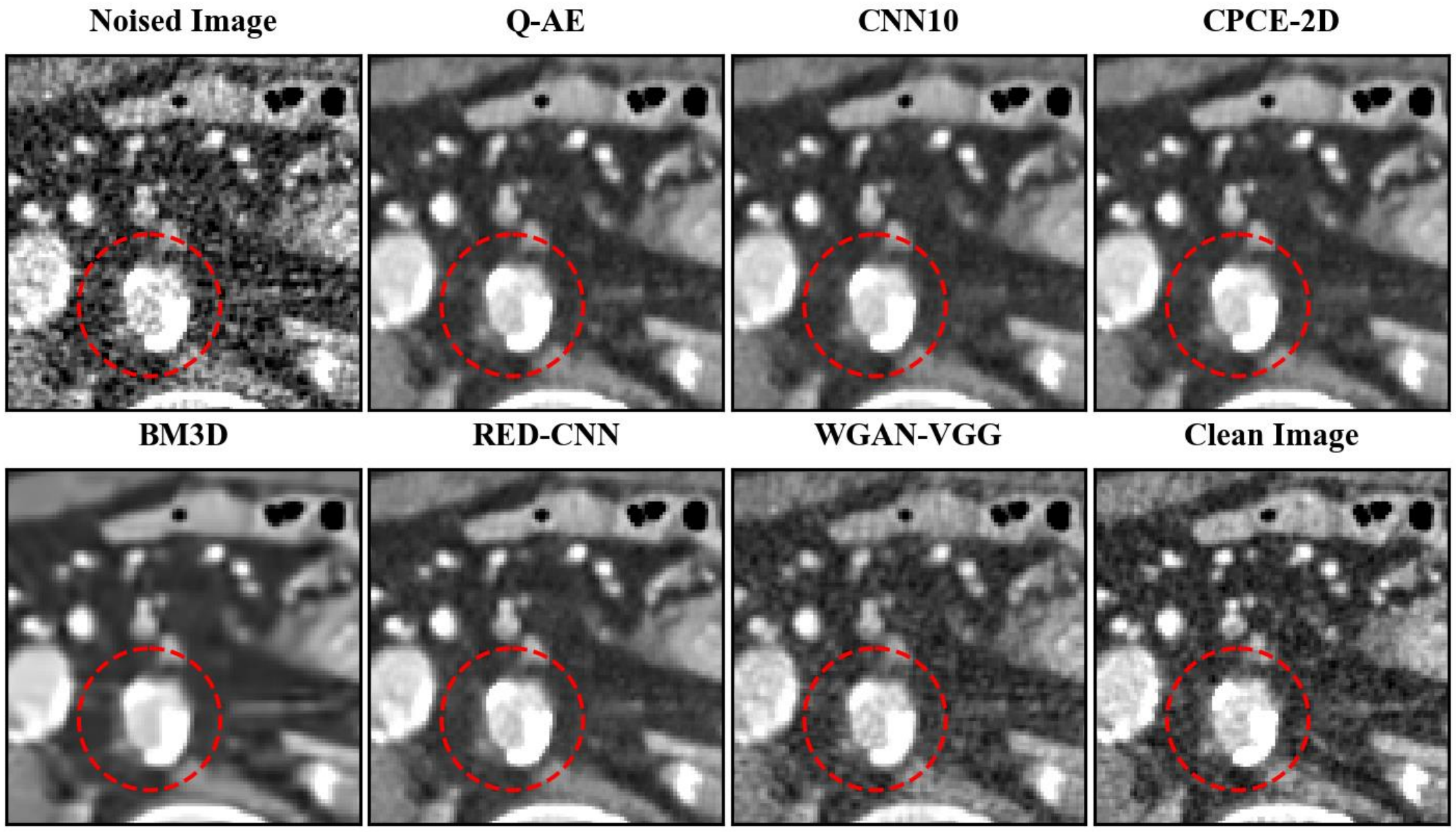

Figure 5. Zoomed ROI of Figure 4. Q-AE image has lower image noise and superior fidelity of small structures. Display window = [-160, 240].

Furthermore, as shown in Figures 4-7, our radiologist coauthors concluded that Q-AE is visually slightly superior in terms of noise removal and detail preservation, although a light gap can be observed in the spinal part in the clean image while it disappeared in the Q-AE result. Specifically, compared to the other techniques, the Q-AE images contained lower image noise, maintained conspicuity of small organism (in the small red circle) and superior fidelity of small structures.

In addition to qualitative analysis, we computed three image quality metrics for quantitative comparison: peak-to-noise ratio (PSNR), structural similarity (SSIM), and root mean square error (RMSE). To reveal the potential of Q-AE comprehensively and objectively in comparison to other SOTA models, we calculated the quantitative metrics on all slices of five patients. The results are presented in TABLE III. It can be seen that the performance metrics of WGAN-VGG, CNN10 and BM3D are all inferior to that of Q-AE and RED-CNN. Generally speaking, Q-AE and RED-CNN perform comparably but Q-AE merely employs no more than one fourth of the parameters in RED-CNN. In terms of computational cost for a forward pass, the number of GFLOPs ($10^9$ float operations) of RED-CNN is four times that of Q-AE. The number of parameters, GFLOPs and inference time are all shown in TABLE V, the best scores are bolded therein.

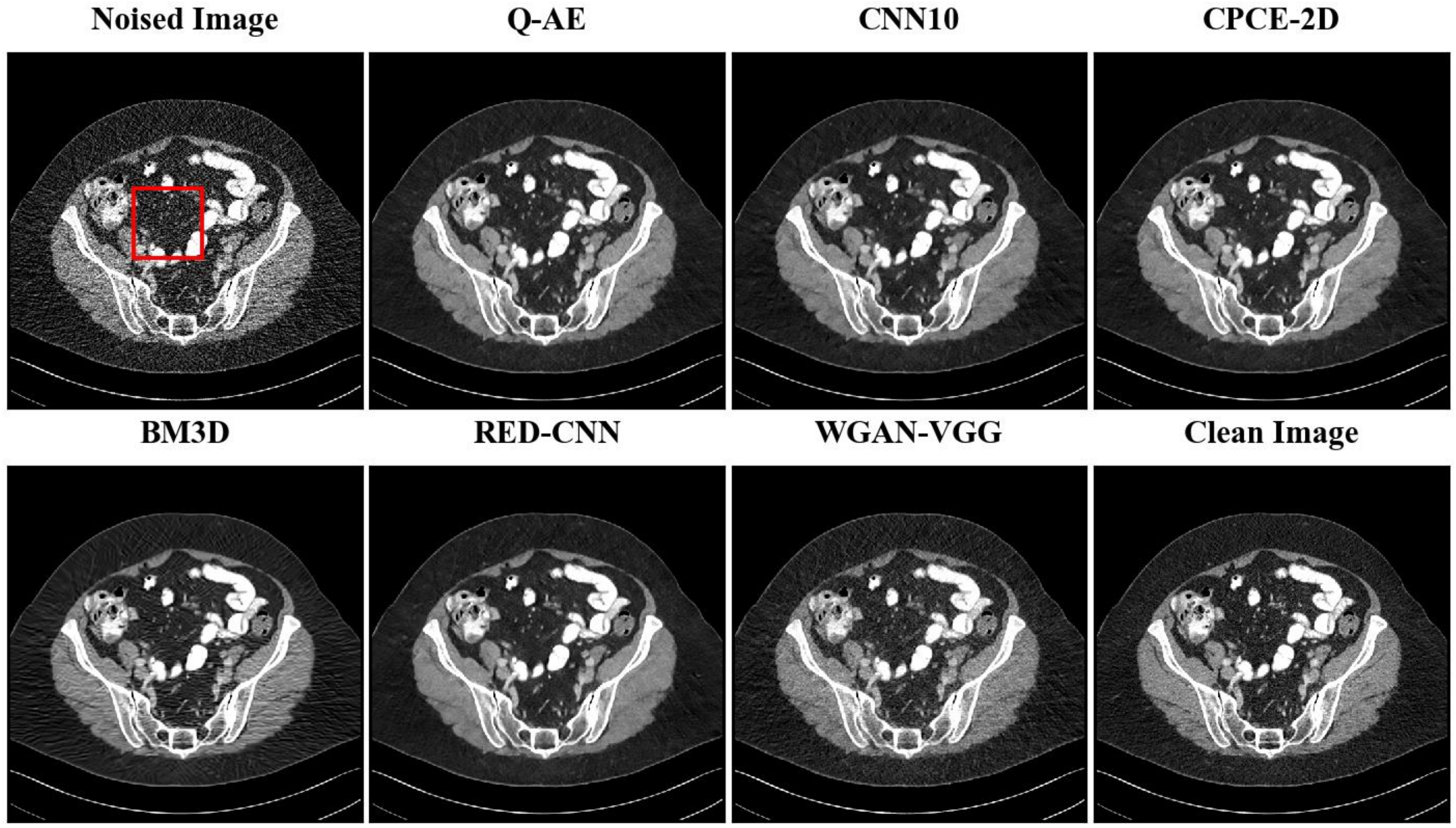

Figure 6. Comparison of different deep learning methods for denoising low dose CT images. Compared to the other techniques, the Q-AE image has lower image noise, maintained conspicuity of vascular calcifications and small lymph nodes (in the small red circle), and superior fidelity of small structures. The display window = [-160, 240].

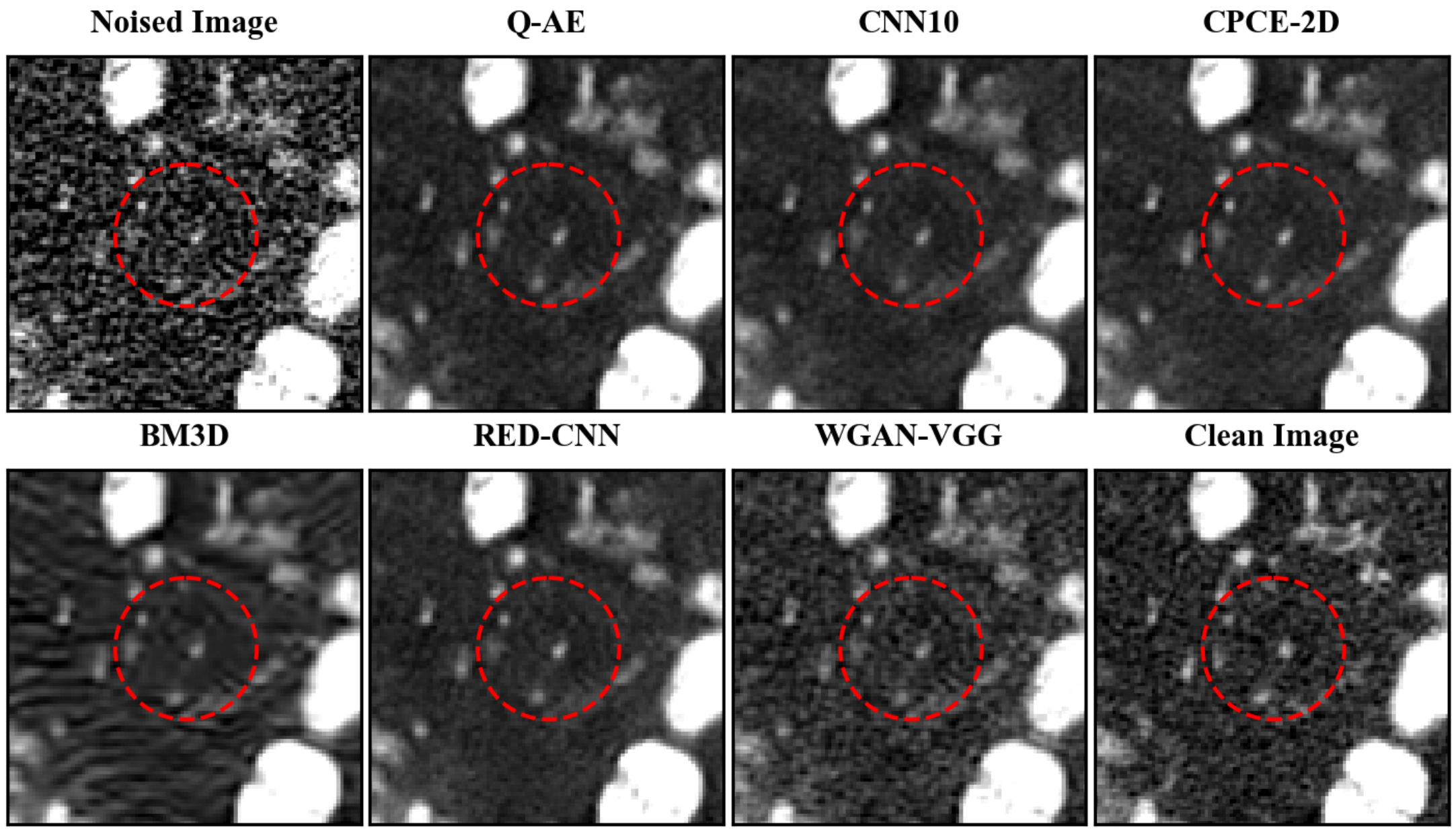

Figure 7. Zoomed ROI of Figure 6. Q-AE image has lower image noise and superior fidelity of small structures. The display window = [-160, 240].

In the low-dose CT denoising task, there are three aspects to be emphasized: noise removal, texture preservation, and structure fidelity. Algorithms are supposed to demonstrate a reasonable balance between these three demands in order to help radiologists with their clinical diagnosis. Generally, radiologists have their own criteria to evaluate the quality of restored images. To convincingly testify if Q-AE performs well in the clinical sense, we prepared 15 de-identified cases where noised images (low dose), clean images (normal dose), and images denoised with the six algorithms for a reader study by three experienced radiologists (R1: 19 years; R2: 18 years; R3: 5 years). The three radiologists independently evaluated all 8 JPEG single image series to evaluate image texture, image noise, and fidelity of small structures. Each aspect was assigned a score of 1-4 (1= Unacceptable for diagnostic interpretation, 2= Suboptimal, acceptable for limited diagnostic interpretation only, 3= Average, acceptable for diagnostic interpretation, 4= Better than usual, acceptable for diagnostic interpretation). The CT findings detected were colonic diverticulosis, intrahepatic biliary dilation, focal hepatic lesions, ileostomy with hernia, cystic lesion / fluid collection in pelvis, left ureteric calculus, bladder wall thickening, pelvic lipomatosis, peritoneal nodule, intramuscular lipoma, right rectus muscle atrophy intrathecal focal lesion, and arterial calcifications. The median scores for the images assessed are presented in TABLE IV.

Compared to the low-dose CT images, all the algorithms achieved certain improvements in some aspect. Because the purpose of these algorithms is to denoise, their improvement in noise suppression is most significant. The results of CNN10 and CPCE-2D are comparable in terms of the three measures. RED-CNN and Q-AE perform much better than CNN10, CPCE-2D and WGAN-VGG. Interestingly, Q-AE and CPCE-2D perform equally well in noise reduction and fidelity. But in terms of image texture, Q-AE is slightly better. Overall, Q-AE works most desirably according to the radiologists' evaluation.

### *5). Model Efficiency*

Model efficiency is a major concern in deep learning. As we postulated earlier, although the number of parameters for a quadratic neuron is three times that of a conventional neuron, the complexity of the quadratic network structure can be reduced without compromising its performance. The two theorems described earlier is in support of our hypothesis. Based on these findings and insights, we compared the number of trainable variables of the above methods to demonstrate the model efficiency of Q-AE, as shown in TABLE V. The trainable parameters are those parameters that are updated during the backpropagation. Clearly, Q-AE is the most compact one, and yet its denoising performance is still superior to other deep learning methods. RED-CNN has an almost four times complexity than Q-AE but still performs second to Q-AE. Again, it roughly holds true that the complexity of network equals to the complexity of individual neurons times the structural complexity. While the complexity of the proposed quadratic function is three times that of a conventional inner product, Q-AE only employs 15 $3*3$ filters (225 $3*3$ kernels) in each layer while RED-CNN has 32 $5*5$ filters (1024 $5*5$ kernels) in each layer, which means that RED-CNN has four times more parameters than Q-AE. WGAN-VGG employs a large number of parameters because it stacks extra perceptual modules as part of the network.

We compared the GFLOPs and the inference times of the selected models in a forward pass, where GFLOPs stands for $10^9$ float operations per second, and the inference time is for processing the same $512*512$ low-dose CT image. The comparative results are shown in TABLE V. WGAN-VGG demands the most GFLOPs because it is complicated, involving a generator, a discriminator, and a VGG which enables the perceptual loss. RED-CNN takes the second place in terms of GFLOPs, while CNN10, CPCE-2D and Q-AE have comparable GFLOP costs. As far as the inference time is concerned, CNN10 has the lowest inference time, surprisingly followed by WGAN-VGG since the inference is only implemented by the generator that has a compact structure. In contrast, as indicated by GFLOPs, RED-CNN and Q-AE have higher inference times than the rest. Overall, Q-AE has the lowest number of parameters but at an additional computational cost.

TABLE V: NUMBERS OF TRAINABLE PARAMETERS, FLOPs, INFERENCE TIMES USED IN THE COMPARED MODELS

| Methods | # of Parameters | FLOPs (G) | Inference Times (s) |
|---|---|---|---|
| **CNN10** | 55,872 | 81.84G | 0.5563 |
| **RED-CNN** | 206,400 | 289.25G | 1.9088 |
| **WGAN-VGG** | 21,488,705 | 1320.92G | 0.5822 |
| **CPCE-2D** | 62,016 | **29.70G** | 0.7396 |
| **Q-AE** | **49,818** | 72.87G | **1.2945** |

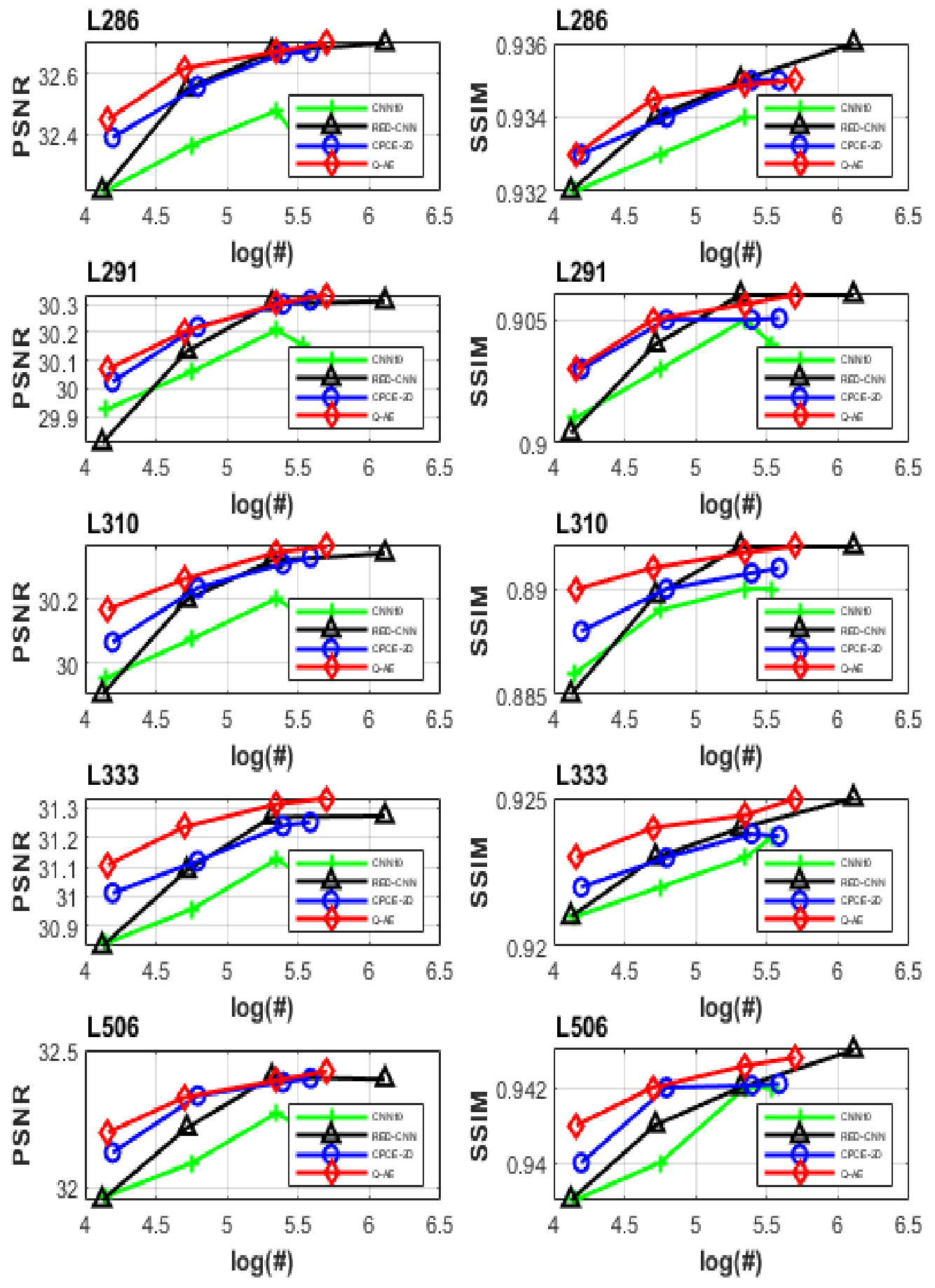

Figure 8. Relationships between numbers of trainable parameters and associated PSNR values for the four models (CNN10, Q-AE, RED-CNN and CPCE-2D). The x axis is made on the log10 scale.

To study the model efficiency and denoising ability of Q-AE, we analyzed the relationships between numbers of trainable parameters in the four models (CNN10, Q-AE, RED-CNN and CPCE-2D) and associated outcomes. WGAN-VGG was excluded because it has enormously many parameters relative to the selected models, and also the main novelty of WGAN-VGG is in the loss function utilized, instead of the network structure. One model with high model efficiency can scale up better even with a less number of parameters. The structures of four models and corresponding numbers of trainable parameters are in TABLE VI. The structures vary in terms of the number of channels used in each layer. We used the original model designs with the identical numbers of channels for each layer. It can be observed that the selection of parameters renders the four models have similar complexities except for RED-CNN(80), where the number denotes the number of channels (this notation is used throughout the manuscript).

TABLE VI: STRUCTURE OF CPCE-2D, CNN10, Q-AE, RED-CNN AND CORRESPONDING NUMBERS OF TRAINABLE PARAMETERS

| | | | | | |
|---|---|---|---|---|---|
| **CNN10** | Channels | 16 | 32 | 64 | 80 |
| | # of Parameters | 14112 | 55872 | 222336 | 347040 |
| **Q-AE** | Channels | 8 | 15 | 32 | 48 |
| | # of Parameters | 14475 | 49818 | 223779 | 501555 |
| **RED-CNN** | Channels | 8 | 16 | 32 | 80 |
| | # of Parameters | 13200 | 52000 | 206400 | 1284000 |
| **CPCE-2D** | Channels | 16 | 32 | 64 | 80 |
| | # of Parameters | 15648 | 62016 | 246912 | 385400 |

We used the same hyper-parameters for all the models as before and calculated the mean PSNR and SSIM values achieved by different model structures on all the slices with respect to five patients (labeled as L286, L291, L310, L333 and L506). Results are shown in Figure 8, where the x axis is made on the log10 scale for visualization. There are several points to be underlined in Figure 8. First, for all the five patients, the PSNR and SSIM plots of Q-AE are above the plots for the other models in the upper left side, which means that Q-AE achieved the highest model efficiency among all the models. Second, as far as PSNR is concerned all the best performance metrics are from Q-AE(48), despite that RED-CNN(64) that used a doubled number of parameters. As for SSIM, RED-CNN(64) is among the best, followed by Q-AE(48). The third point lies in the performance difference among those compact models. When the number of parameters is between 15,000 and 50,000, the PSNR and SSIM measures of Q-AE are larger than that of the other models with more parameters, which suggests that Q-AE is more desirable for compact modeling. Therefore, we conclude that in terms of both the model efficiency and denoising performance, Q-AE is competitive compared to other SOTA algorithms.

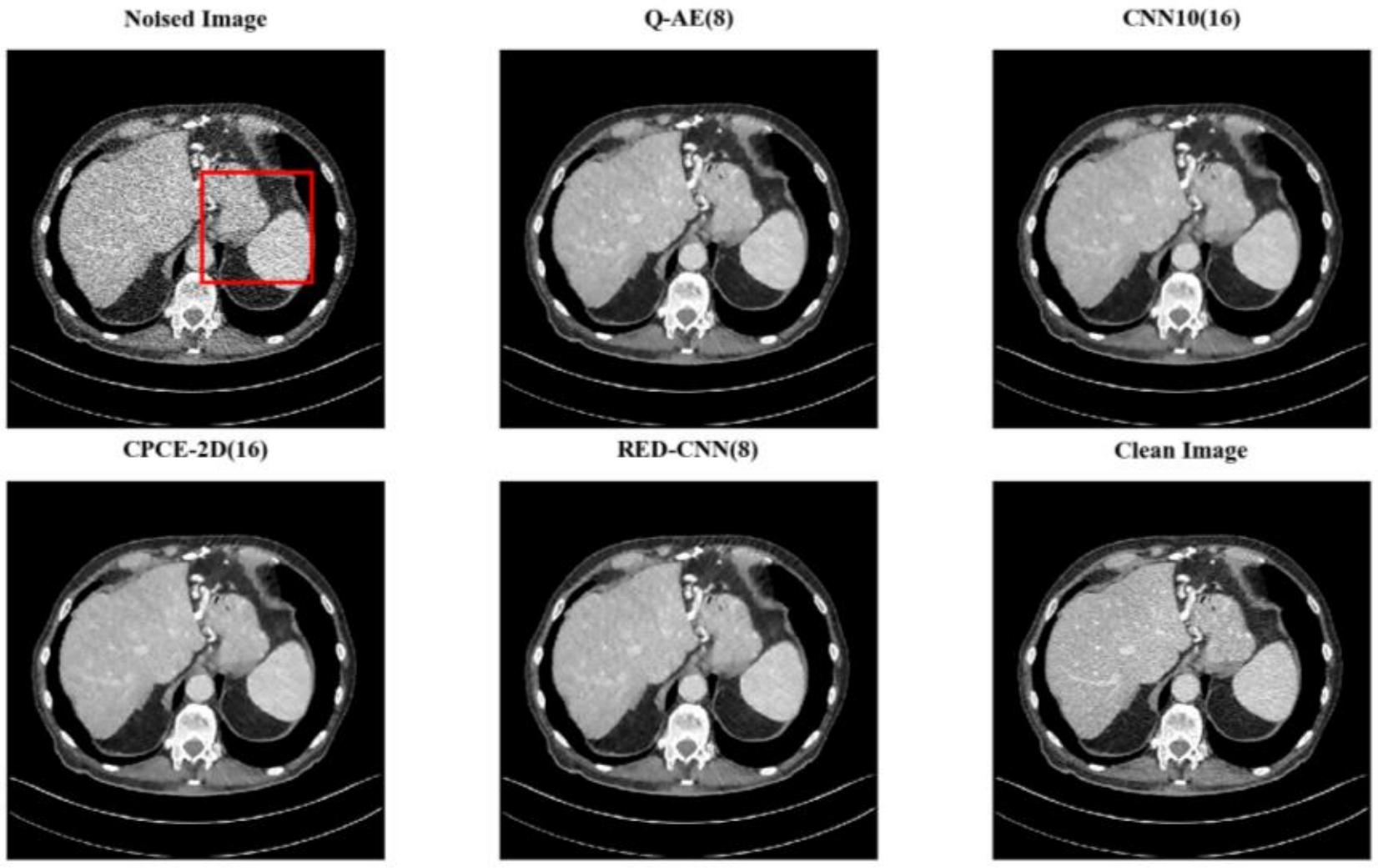

Figure 9. Comparison of different deep learning methods for low-dose CT images. The lesions are highlighted with higher contrast in Q-AE image. Display window = [-160, 240].

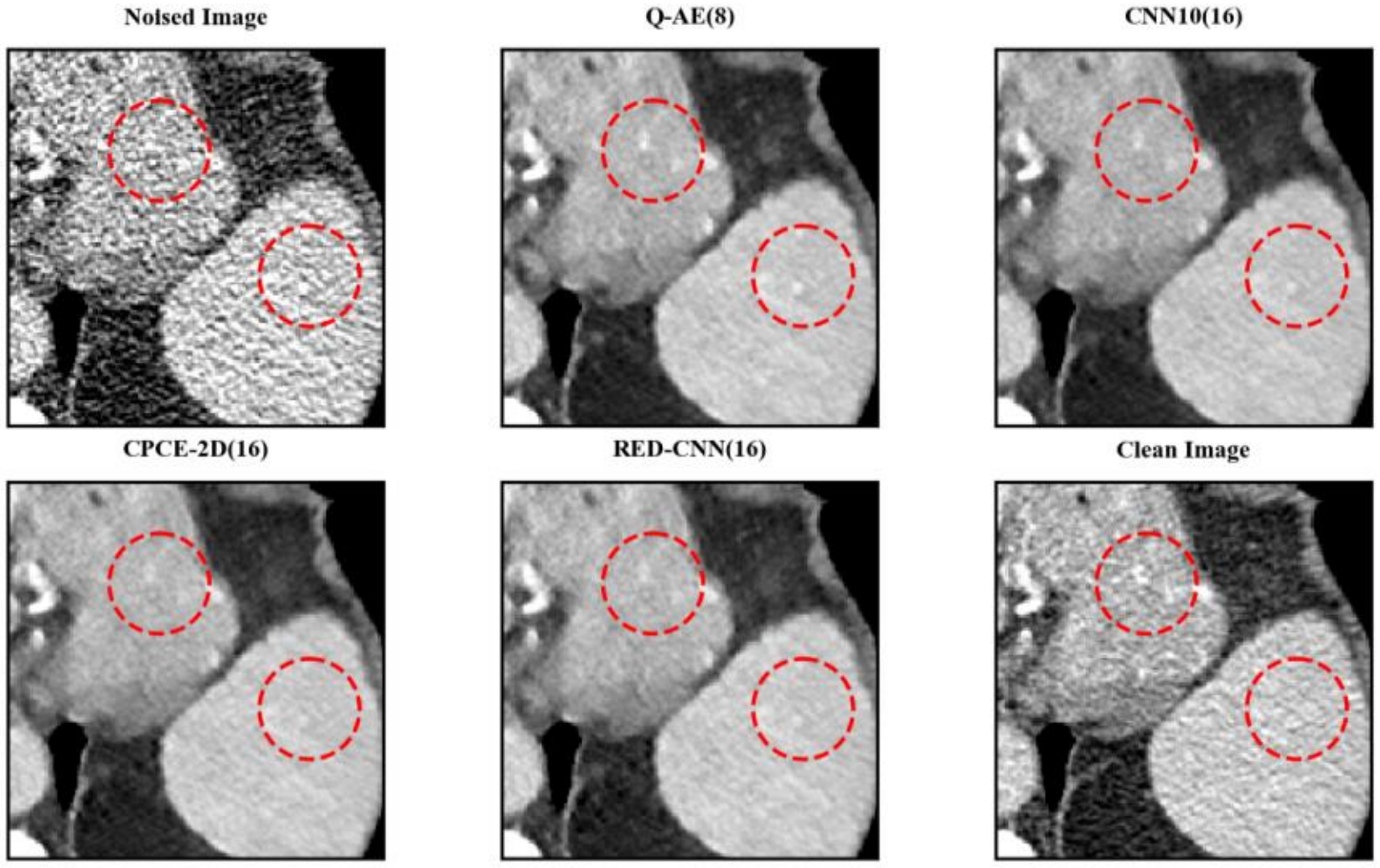

Figure 10. Zoomed ROI of Figure 5.

We utilized one representative slice (50th slice of L506) to visualize the restored results of Q-AE(8), CNN(10), CPCE-2D(16) and RED-CNN(8), which have comparable numbers of parameters, as shown in Figures 9 and 10. It is noticed in Figure 10 that the lesions in the red circles are revealed with a higher contrast using Q-AE(8) than that in the cases using other models.

### *6). Over-fitting*

Over-fitting is a risk in deep learning, referring to the fact that one model may work well on a training dataset but fails to achieve a good performance on a test dataset. Generally, the more complex the model is, the more easily the model tends to over-fit [48]. In the preceding subsection, we have shown that Q-AE enjoys the lowest model complexity, hence Q-AE is less likely to over-fit. To back this proposition numerically, we plot the training loss curve vs validation loss curve in Figure 11. Clearly, as the training loss goes down, the validation loss decreases as well. Their trajectories are similar in the later training stage. According to this observation, it is concluded that Q-AE is appropriately fitting the low-dose CT denoising task.

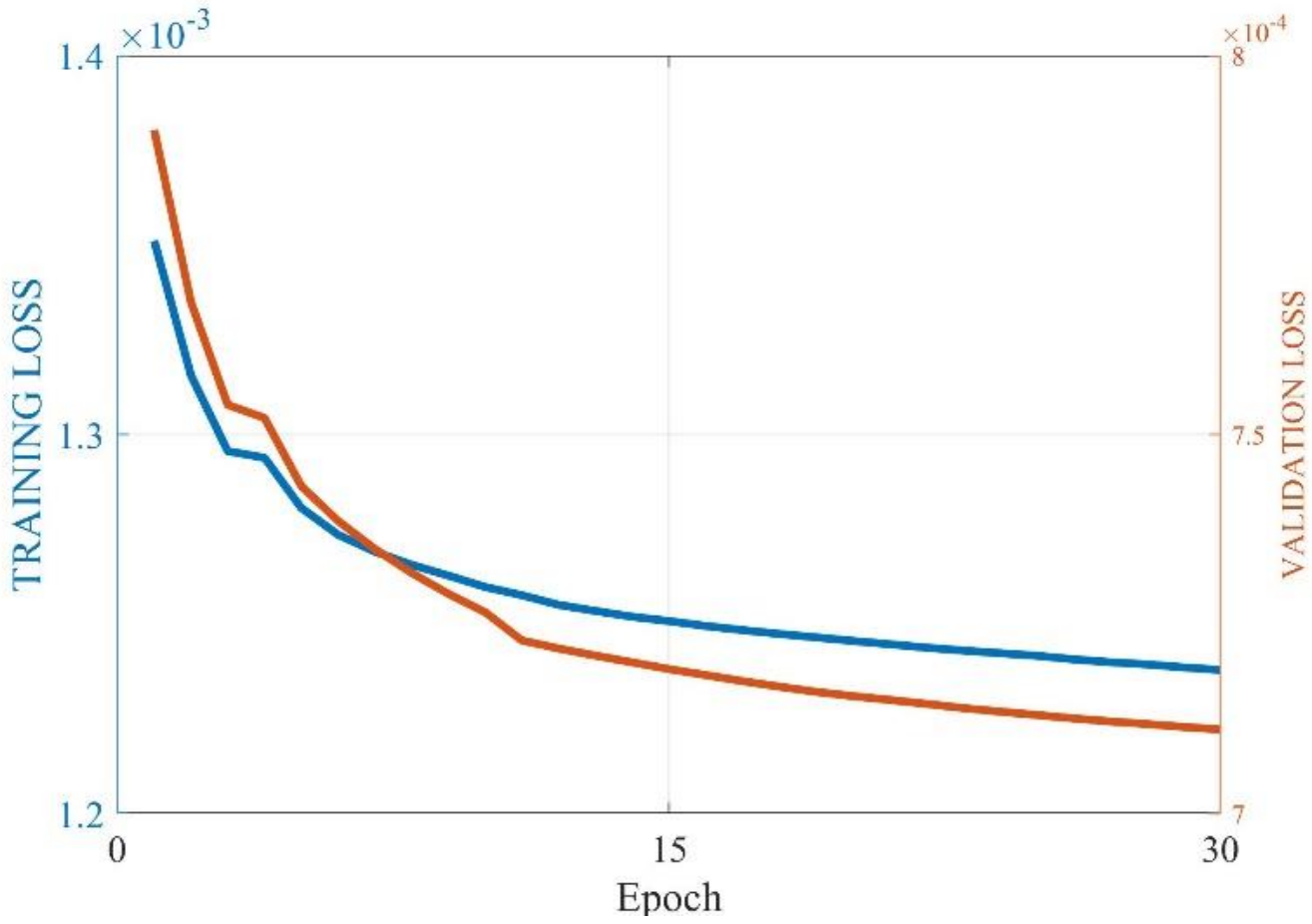

Figure 11. Descent trend of the training loss is consistent to that of the validation loss for Q-AE.

### *7). Training by Transfer Learning*

Conventionally, to train quadratic networks we can either randomly initialize all the parameters or deploy transfer learning. Transfer learning [49] is a type of machine learning techniques with which a model developed for a task can be adapted for another related task. Weight transfer is a typical transfer learning technique, which initializes a new model with the parameters of a pre-trained model. Since a conventional neuron is a special case of a quadratic neuron, it is feasible to initialize a quadratic model by a conventional model of the same structure. Specifically, suppose that we have $\boldsymbol{w}^{\boldsymbol{1^{st}}}, \boldsymbol{b}^{\boldsymbol{1^{st}}}$ of a trained conventional neuron, then the parameters of the corresponding quadratic neuron can be initialized in the following way:

$$\begin{cases} \boldsymbol{w_r} = \boldsymbol{w}^{\boldsymbol{1^{st}}} \quad \boldsymbol{w_g} = \boldsymbol{1} \\ \boldsymbol{b_r} = \boldsymbol{b}^{\boldsymbol{1^{st}}}, \quad \boldsymbol{b_g} = 1, \quad \boldsymbol{c} = \boldsymbol{0} \end{cases} \tag{9}$$

Note that $\boldsymbol{w_b}$ is still randomly initialized, because $\boldsymbol{w_b x^2}$ is not in the conventional neuron.

In contrast, our initialization strategy is to inhibit the quadratic terms by setting $\boldsymbol{w_g} = \boldsymbol{0}, \boldsymbol{b_g} = \boldsymbol{1}$ and $\boldsymbol{w_b}$ with a small variance in the beginning and let the model itself optimize the quadratic terms adaptively. To test the effectiveness of our strategy, we compared its convergence behavior with that of transfer learning and training from scratch. We initialized $\boldsymbol{w_b}$ in every layer using constant initialization with 0, 0.001 and 0.003 respectively. Meanwhile, we follow the earlier initialization strategy for training from scratch. Because pervious experiments show that Q-AE has been stable in the last 10 epochs, we use 20 epochs in this experiment. The denoising results were evaluated at each epoch on the validation dataset, as shown in Figure 12. Interestingly, the effectiveness of transfer learning *per se* is impacted by how $\boldsymbol{w_b}$ is initialized. When $\boldsymbol{w_b}$ is initialized with a constant 0, the transfer learning strategy ends up with a higher validation loss. When $\boldsymbol{w_b}$ is initialized with a higher value, transfer learning can beat the randomized initialization. One highlight is that although transfer learning provides a good initial guess for Q-AE, overall, transfer learning and our strategy converge to the same level of validation loss. which is actually good news because our strategy, without a need of training another model, works as well as transfer learning. Please also note that Q-AE trained either way outperforms the conventional network Q-AE($1^{st}$) that shares the same structure with Q-AE.

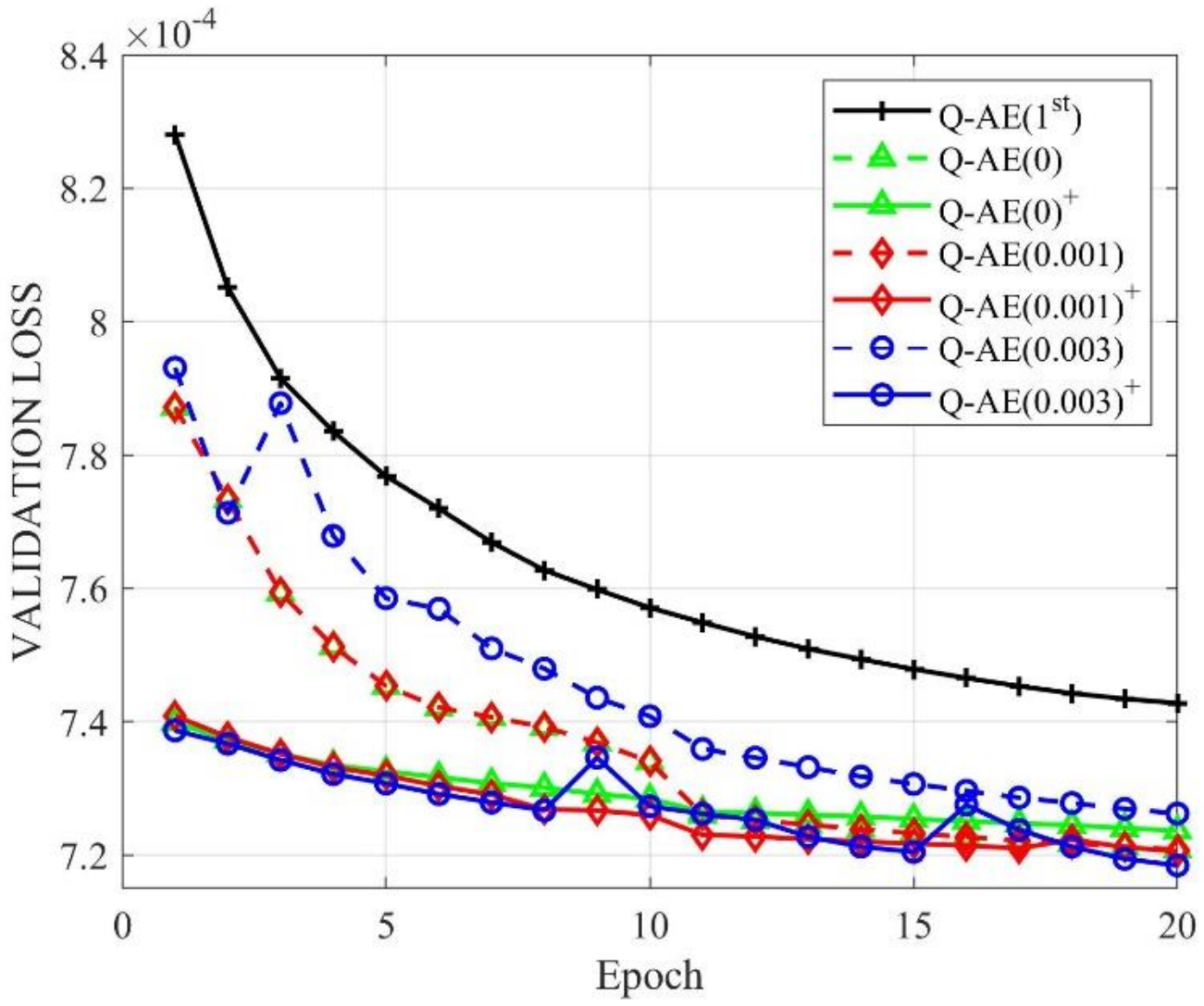

Figure 12. Comparison between training from scratch and training via weight transfer in terms of the validation loss. Q-AE($1^{st}$) is the conventional network in the same structure of Q-AE. The number x in the legend means $\boldsymbol{w_b}$ is initialized with constant=x. The quadratic model trained via transfer learning is marked by a superscript +.

While it is premature to formulate a general mechanism behind these findings, our intuition is that the weights from the trained conventional network may not reflect the authentic capacity of the quadratic counterpart. because at the cellular level they are different, weights of a conventional network can only initialize the corresponding quadratic network partially, without any influence on the quadratic terms. In other words, how to train a quadratic network optimally remains an open question.

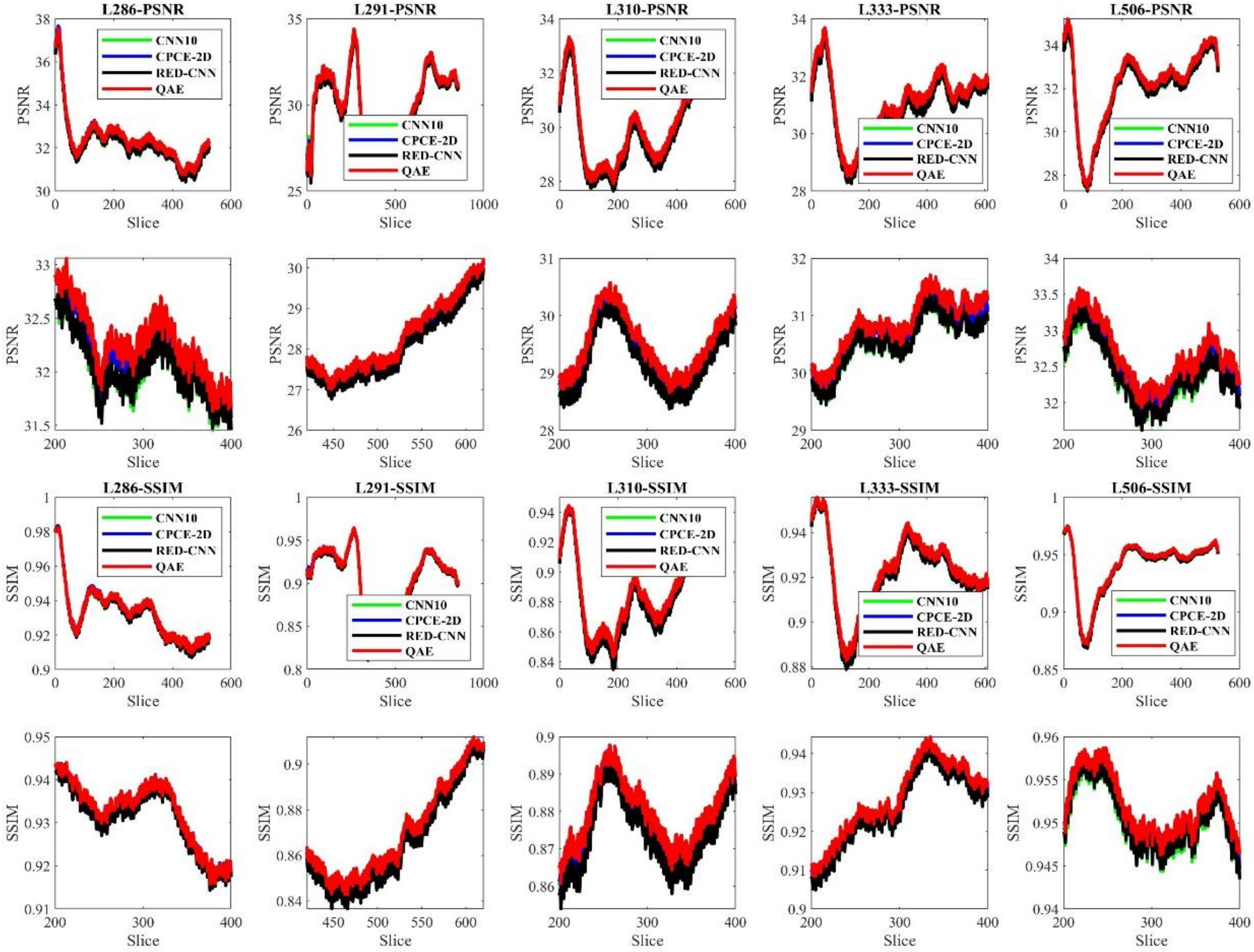

Figure 14. Lower five images are zoomed slices from the whole slices for better visualization. Q-AE consistently performs the best over other three models.

*8). Robustness*

**Robustness to initialization:** we tested the robustness of Q-AE with respect to the random initialization. We repeatedly and randomly initialized $\boldsymbol{w_r}$ of Q-AE with the Gaussian function five times on the conditions of $\boldsymbol{w_b = 0, w_g = 0, b_g = 1, w_b = 0.001, w_g = 0, b_g = 1}$ and $\boldsymbol{w_b = 0.003, w_g = 0, b_g = 1}$ , denoted with blue, red and green respectively in Figure 13. Please note that the above conditions were the same as that for the experiments where we compared our initialization strategy with the transfer learning strategy. As shown in Figure 13, all the models converged eventually to a low loss. Although undergoing severe oscillations earlier for some red curves and blue curves, Q-AE still converged stably.

**Robustness to anatomical slices:** we also evaluated the robustness of Q-AE with respect to anatomical structure. Our curiosity is if Q-AE only performs well just on some slices while doing inferiorly on the other slices. We have respectively tested CNN10(16), Q-AE(8), RED-CNN(8) and CPCE-2D(16) on all the slices of five patients. The results are shown in Figure 12. These four models have comparable numbers of trainable parameters. It is underscored that Q-AE consistently performed superbly in terms of PSNR and SSIM on all the slices among all the algorithms.

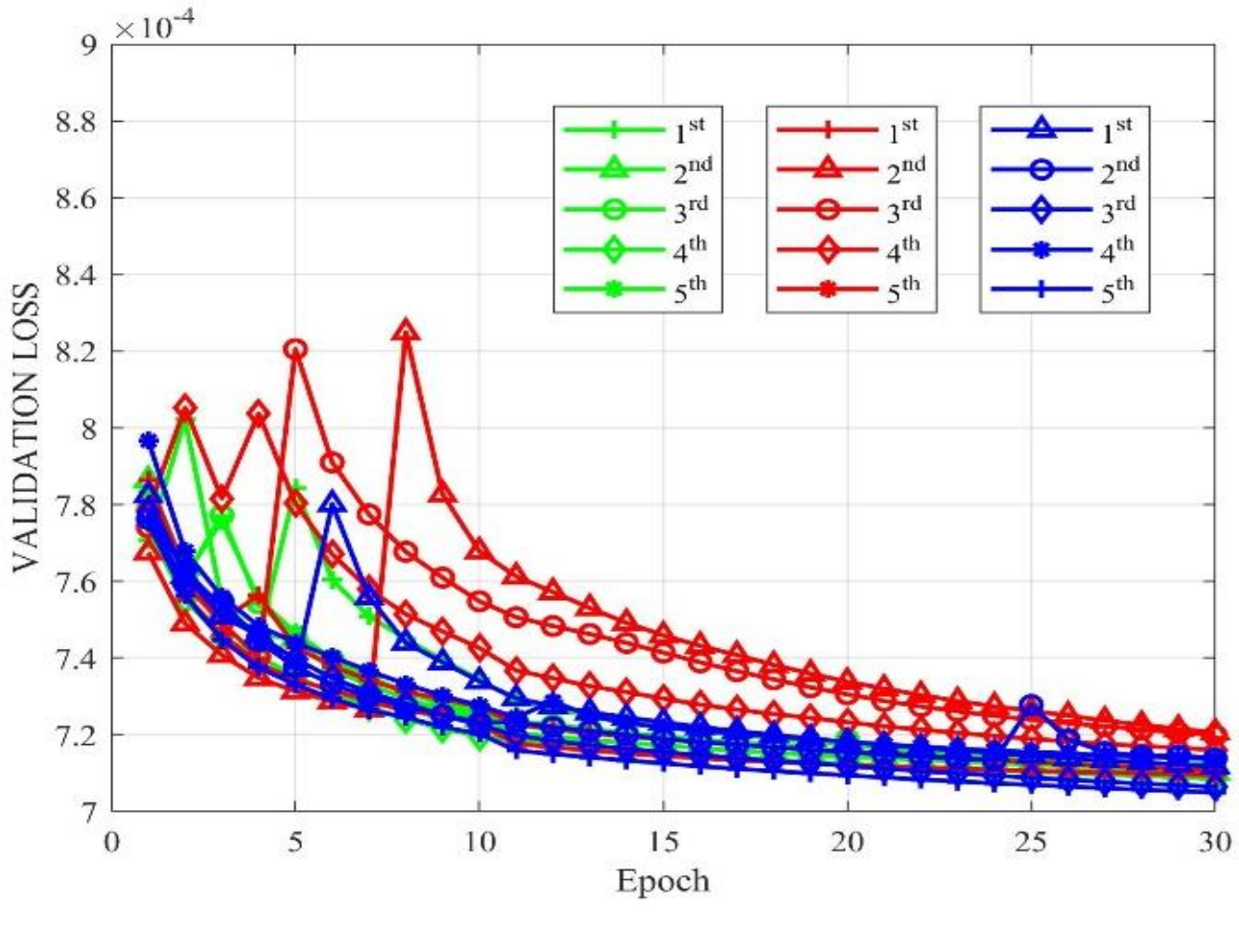

Figure 13. Convergence behavior of Q-AE with different initialization methods. Q-AE converged stably after oscillations in an early stage.

*9). Intrinsic Denoising Ability of Quadratic Networks*

Next, we tested the intrinsic representation ability of quadratic networks. Our curiosity is if the denoising gain of a network increases or decreases after we replace all the quadratic neurons with conventional neurons in this low-dose CT task (by conventional neurons, we mean neurons based on the inner product). We would like to mention that while RED-CNN deploys 5-by-5 kernels, our quadratic network utilizes 3-by-3 kernels, therefore RED-CNN is not the first-order version of our Q-AE topologically. In this experiment, we replaced the quadratic neurons of Q-AE(8), Q-AE(15), Q-AE(32) and Q-AE(48) with conventional neurons respectively, the resultant networks are denoted as AE(8), AE(15), AE(32) and AE(48). We set the optimal learning rate for the AE to $5 * 10^{-4}$. All the other hyperparameters for the training of AEs remained the same as that for Q-AE. After training, we applied the trained networks to all the slices of five patients and computed the mean RMSE values because the loss function used in the training is MSE. The results are shown in Figure 15.

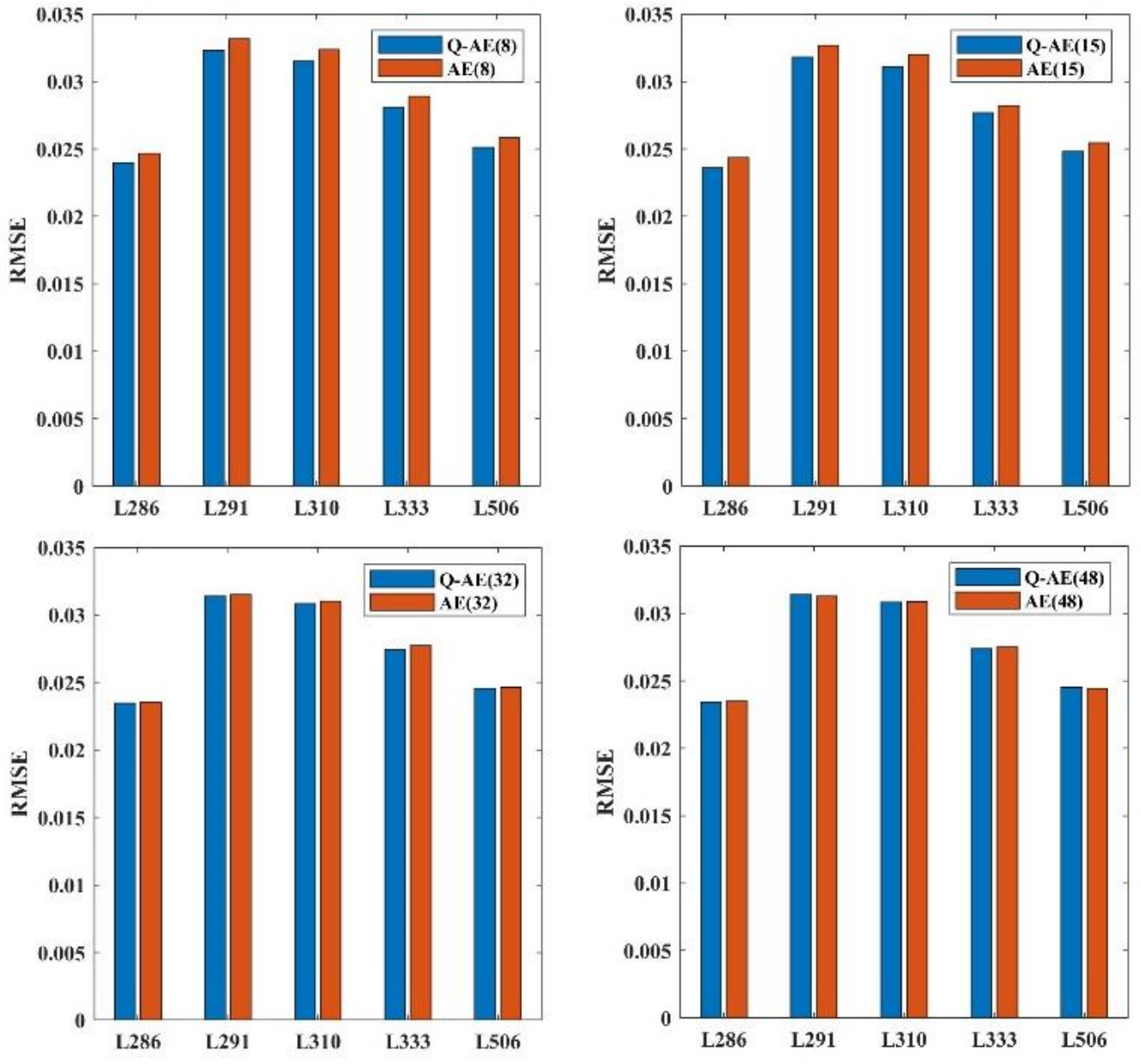

Figure 15. Employment of quadratic neurons can indeed reduce RMSE in the denoising task, although the reduction effect diminishes as the network becomes increasingly complex.

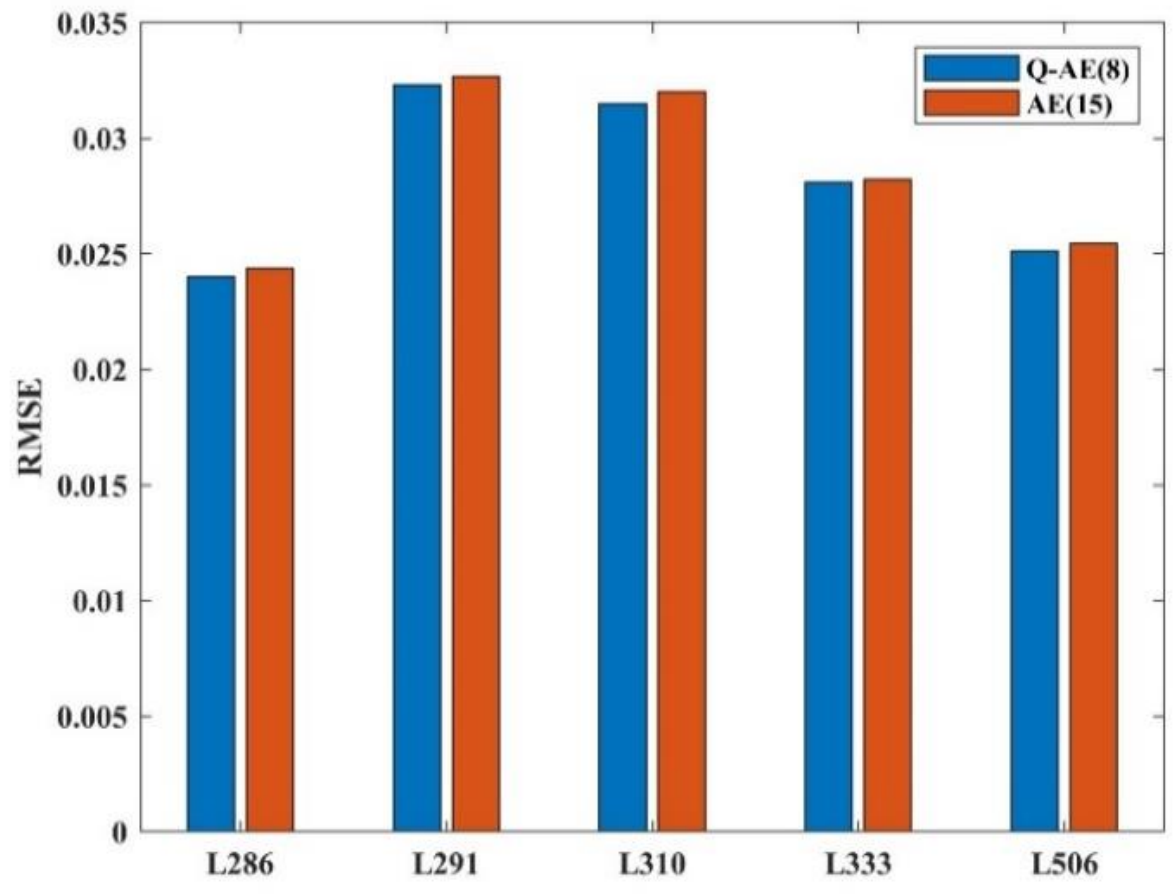

Figure 16. Although the number of Q-AE(8) is less than the number of AE(15), 14475 vs 16470, Q-AE(8) still achieved lower RMSE.

When the network structure is compact, i.e., the numbers of channels are 8 and 15, the RMSE values of Q-AE are 1∼2% lower than those of AEs. When the number of channels went up to 32, the RMSE reduction became 0.5∼1%. When the number of channels was 48, the network structures are sufficiently complicated to saturate the denoising performance, then the difference between quadratic and conventional networks made no significant effect. In summary, the employment of quadratic neurons can indeed lead to RMSE reduction in the denoising task.

Furthermore, we compared Q-AE(8) and AE(15), which have close numbers of trainable parameters, as Figure 16 shown. Although the number of Q-AE(8) is less than the number of AE(15), 14475 vs 16470, Q-AE(8) still achieved lower RMSE in the advantage of 1∼2%.

*10). Comparison with Networks Using Quadratic Activation*

Previously, we argued that there were fundamental differences between our quadratic networks and the networks using quadratic activation. The main reason is that the neurons with quadratic activation are still subjected to linear decision boundaries, which makes a few differences on the representation ability. To justify our argument, we compared our quadratic autoencoder with the autoencoder using quadratic activation. We selected Q-AE(15) as a representative network and replaced the neurons in Q-AE(15) with neurons using quadratic activation. In the literature, there are two kinds of quadratic activations: the standard one, $y = \alpha x^2$ [23] and rectified quadratic activation [52]:

$$y = \begin{cases} \alpha x^2 & if\ x > 0 \\ 0 & if\ x \le 0 \end{cases},$$

where $\alpha$ is the coefficient. In the denoising experiment, $\alpha$ was empirically set as to 0.4. The learning rates for quadratic activation and rectified quadratic activation were set to $5 * 10^{-5}$ and $5 * 10^{-4}$ respectively. All other experimental conditions were kept the same as those used for Q-AE. The convergence behaviors of two models are shown in Figure 17. Unfortunately, the AE with rectified quadratic activation does not converge well, it seemed that the model was trapped in an inferior local minima. We would like to underscore that such fact is not because the learning rate was small since we tried several larger learning rates up to $2 * 10^{-1}$. In contrast, after 30 epochs AE with quadratic activation converged to a reasonable level that was higher than the Q-AE validation loss as shown in Figure 3.

Furthermore, we applied the trained model to the same slices (270th slice from patient L506 and 340th slice from patient L310). Because AE with rectified quadratic activation did not converge well, we only tested the AE with quadratic activation. The denoised results are shown in Figure 18. It can be seen that the images are still noisy with dotty artifacts. We also computed PSNR, SSIM and RMSE of AEs using quadratic activation and Q-AE. The results are summarized in TABLE VII. By each of the three metrics (SSIM, PSNR, RMSE), Q-AE ranked the highest at the two representative slice locations. It is also worth mentioning that RMSE and PSNR scores are consistent to the convergence curves with which Q-AE showed the lower validation loss.

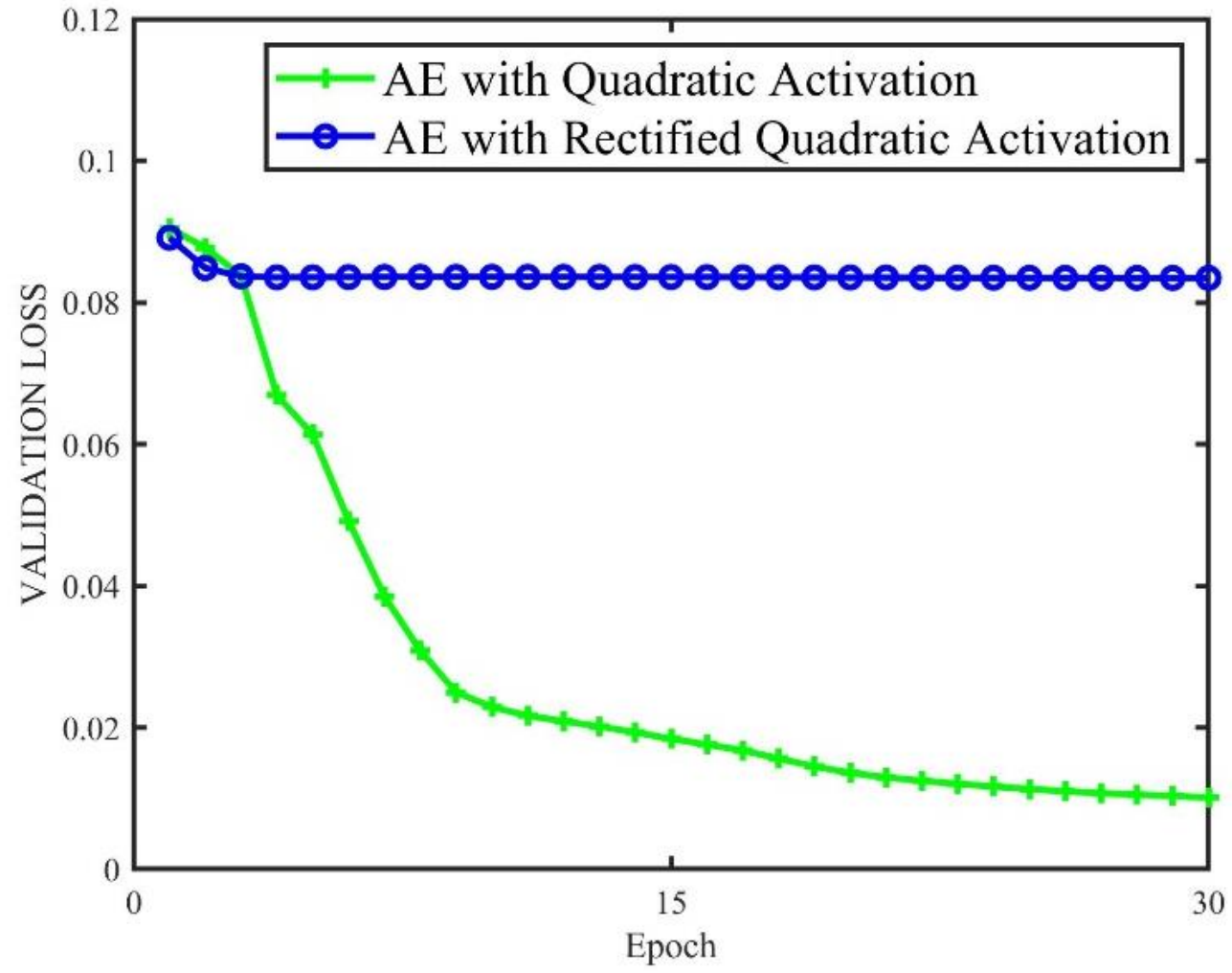

Figure 17. AE with threshold quadratic activation does not converge well, while AE with quadratic activation converged to a level higher than that of Q-AE.

TABLE VII: QUANTITATIVE COMPARISON BETWEEN AE USING QUADRATIC ACTIVATION AND QUADRATIC AUTOENCODER

| | **Fig. 4** | | | **Fig. 6** | | |
|---|---|---|---|---|---|---|
| | **PSNR** | **SSIM** | **RMSE** | **PSNR** | **SSIM** | **RMSE** |
| **LDCT** | 28.230 | 0.90924 | 0.03876 | 23.129 | 0.72137 | 0.06975 |
| **Q-AE** | **32.712** | **0.95266** | **0.02313** | **28.867** | **0.86898** | **0.03602** |
| **AE(QA)** | 31.067 | 0.94113 | 0.02796 | 27.437 | 0.83902 | 0.04248 |

Note: AE(QA) denotes autoencoder using quadratic activation.

## IV. Discussions and Conclusions

We have shown the merits of quadratic deep learning in the context of low-dose CT denoising over state-of-the-art denoising algorithms and hypothesize that this quadratic approach is advantageous in other applications as well, such as radiomics. With quadratic neurons, more complicated features can be extracted to improve the diagnostic performance. For example, with a conventional network for classification, the decision boundary will be piecewise-linearly defined but with a quadratic network the decision boundary will be piecewise quadratic, and much more natural and precise. Currently, all deep learning libraries support the network training via backpropagation through conventional neurons. These training functions should be upgraded to support quadratic neurons. Also, quadratic counterparts of typical network architectures can be prototyped in these libraries so that users have more freedom to adapt and develop. Along this direction, other types of neurons and hybrid networks consisting of diverse types of neurons can be investigated for further development of deep learning techniques. We have shared our code in https://github.com/FengleiFan/Q-AE.

In conclusion, this paper is the first reporting a quadratic deep learning model for medical imaging. Specifically, we have demonstrated that a quadratic autoencoder achieved the best performance numerically and clinically yet with the lowest complexity among the compared models. Since network-based denoising methods are fairly mature, the established superiority of our proposed Q-AE over the state-of-the-art low-dose CT denoising networks is very encouraging. As follow-up projects, more quadratic models will be prototyped and applied to real-world applications.

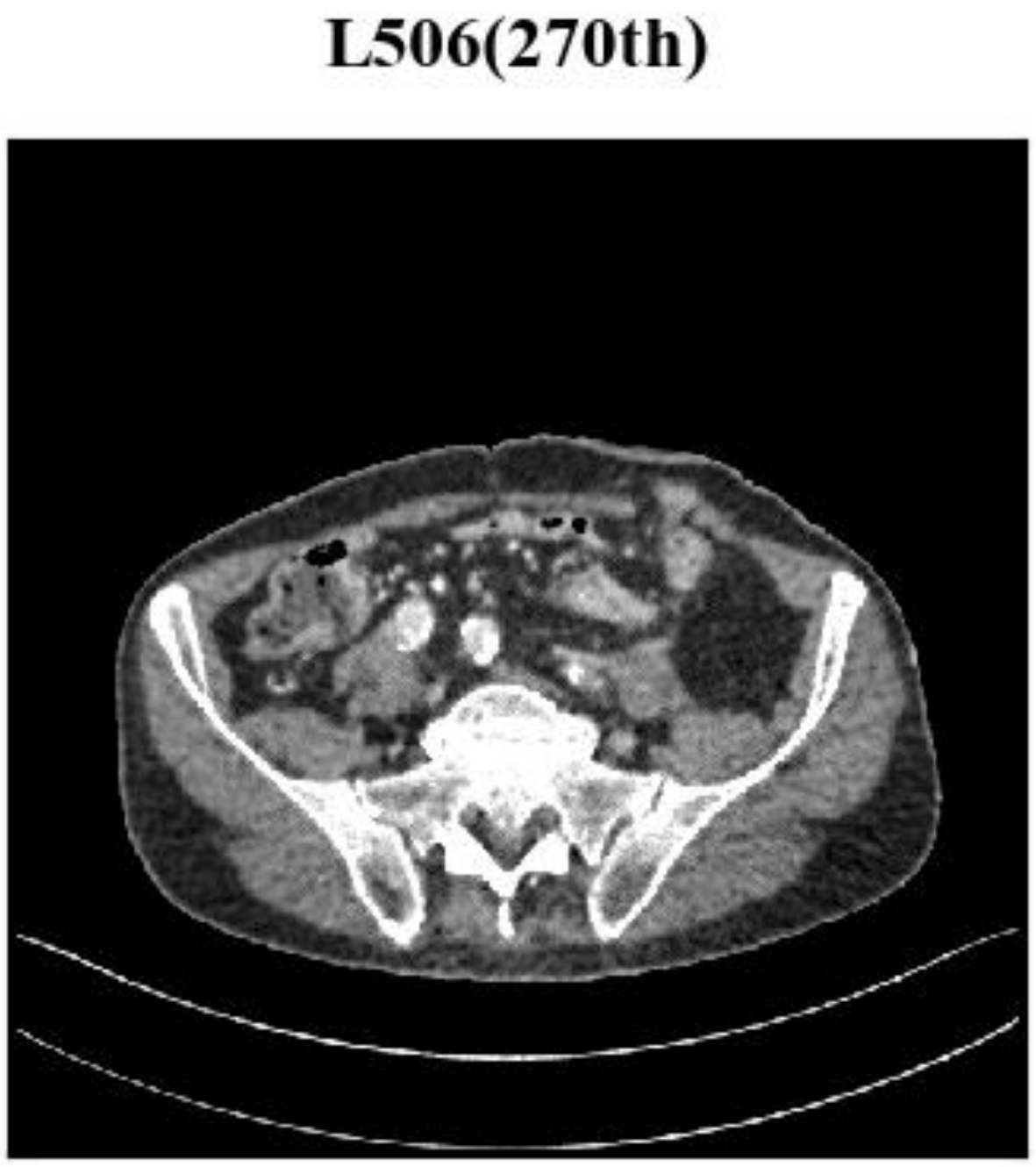

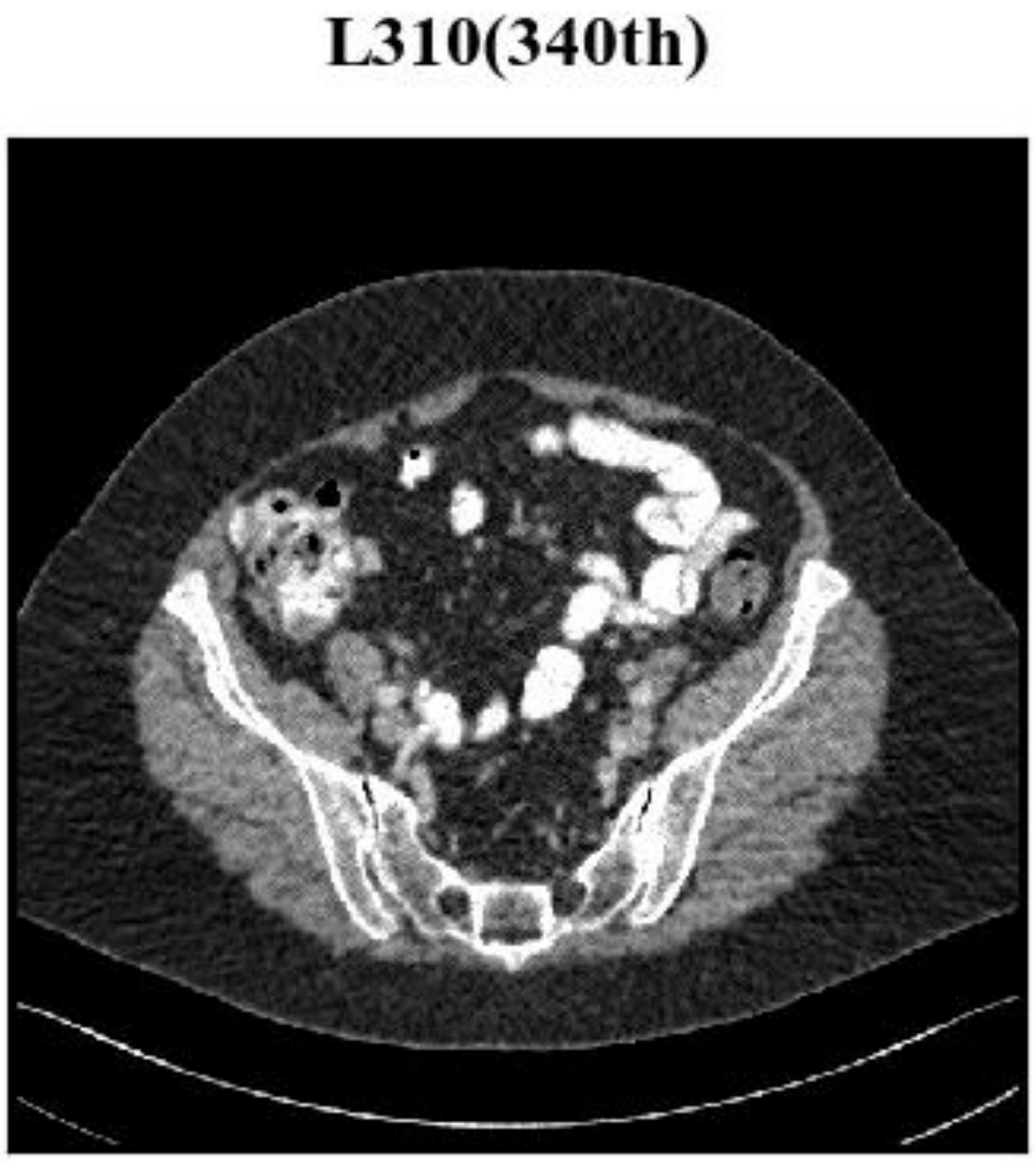

Figure 18. Results from AE using quadratic activation are still noisy with dotty artifacts.